%% file: 0-main.tex
\RequirePackage{fix-cm}
\documentclass[smallextended]{svjour3}       
\smartqed  
\usepackage{graphicx}
\usepackage{times}
\usepackage[numbers]{natbib}
\usepackage{multicol}
\usepackage[bookmarks=true]{hyperref}
\usepackage{xspace}
\usepackage{caption}
\def\tablename{Table}
\usepackage{url}
\usepackage{amsmath,amssymb}
\usepackage{graphicx,color}
\usepackage{verbatim}
\usepackage{gensymb}
\usepackage{setspace}
\graphicspath{{figures/}}

\newcommand{\dold}{Fabric-Random\xspace} 
\newcommand{\dnew}{Fabric-CornerBias\xspace} 
\newcommand{\vsf}{VisuoSpatial Foresight\xspace} 
\newcommand{\il}{Imitation Learning\xspace}
\newcommand{\vsfrss}{VSF-1.0\xspace}
\newcommand{\vsfnew}{VSF-2.0\xspace}

\newcommand{\ba}{\mathbf{a}}

\newcommand{\bo}{\mathbf{o}}

\newcommand{\bz}{\mathbf{z}}

\usepackage{color}
\usepackage[usenames,dvipsnames,table,xcdraw]{xcolor}

\begin{document}

\title{VisuoSpatial Foresight for Physical Sequential Fabric Manipulation}




\titlerunning{VisuoSpatial Foresight for Physical Sequential Fabric Manipulation}

\author{Ryan Hoque* \and Daniel Seita* \and Ashwin Balakrishna \and Aditya Ganapathi \and Ajay Kumar Tanwani \and Nawid Jamali \and Katsu Yamane \and Soshi Iba \and Ken Goldberg
\thanks{$^{*}$ Equal contribution}%
}

\authorrunning{Ryan Hoque*, Daniel Seita* et al.} 

\institute{Ryan Hoque \at
              ryanhoque@berkeley.edu \\
           \and
           Daniel Seita \at
              seita@berkeley.edu
}

\date{}

\maketitle

\begin{abstract}
Robotic fabric manipulation has applications in home robotics, textiles, senior care and surgery. Existing fabric manipulation techniques, however, are designed for specific tasks, making it difficult to generalize across different but related tasks. We build upon the Visual Foresight framework to learn fabric dynamics that can be efficiently reused to accomplish different sequential fabric manipulation tasks with a single goal-conditioned policy. We extend our earlier work on VisuoSpatial Foresight (VSF), which learns visual dynamics on domain randomized RGB images and depth maps simultaneously and completely in simulation. In this earlier work, we evaluated VSF on multi-step fabric smoothing and folding tasks against 5 baseline methods in simulation and on the da Vinci Research Kit (dVRK) surgical robot without any demonstrations at train or test time. A key finding was that depth sensing significantly improves performance: RGBD data yields an $\mathbf{80 \%}$ improvement in fabric folding success rate in simulation over pure RGB data. In this work, we vary 4 components of VSF, including data generation, visual dynamics model, cost function, and optimization procedure. Results suggest that training visual dynamics models using longer, corner-based actions can improve the efficiency of fabric folding by 76\% and enable a physical sequential fabric folding task that VSF could not previously perform with 90\% reliability. Code, data, videos, and supplementary material are available at \url{https://sites.google.com/view/fabric-vsf/}.
\end{abstract}

\input{1-introduction.tex}
\input{2-related-work.tex}

\input{3-prob-statement.tex}
\input{4-approach.tex}
\input{4.5-implementation.tex}
\input{5-simulated-exps.tex}
\input{6-physical-exps.tex}
\input{7-conclusion.tex}

\begin{acknowledgements}
This research was performed at the AUTOLAB at UC Berkeley in affiliation with Honda Research Institute USA, the Berkeley AI Research (BAIR) Lab, Berkeley Deep Drive (BDD), the Real-Time Intelligent Secure Execution (RISE) Lab, and the CITRIS ``People and Robots'' (CPAR) Initiative, and by the Scalable Collaborative Human-Robot Learning (SCHooL) Project, NSF National Robotics Initiative Award 1734633. The authors were supported in part by Siemens, Google, Amazon Robotics, Toyota Research Institute, Autodesk, ABB, Samsung, Knapp, Loccioni, Intel, Comcast, Cisco, Hewlett-Packard, PhotoNeo, NVidia, and Intuitive Surgical. Daniel Seita is supported by a Graduate Fellowships for STEM Diversity and Ashwin Balakrishna is supported by an NSF GRFP. We thank Mohammad Babaeizadeh for advice on extending the SVG model to be action-conditioned, and we thank Ellen Novoseller and Lawrence Chen for extensive writing advice.
\end{acknowledgements}

\section*{Conflict of interest}

The authors declare that they have no conflict of interest.

\bibliographystyle{spbasic}      
\bibliography{vsf}   

\input{8-appendix.tex}

\end{document}

%% file: 1-introduction.tex
\section{Introduction}
\label{sec:intro}
Advances in robotic manipulation of deformable objects has lagged behind work on rigid objects due to the far more complex dynamics and configuration space. Fabric manipulation in particular has applications ranging from senior care~\cite{personalized_dressing_2016}, sewing~\cite{sewing_2012}, ironing~\cite{ironing_2016}, bed-making~\cite{seita-bedmaking} and laundry folding~\cite{laundry2012,folding_iros_2015,folding_2017,shibata2012trajectory} to manufacturing upholstery~\cite{Torgerson1987VisionGR} and handling surgical gauze~\cite{thananjeyan2017multilateral}. However, prior work in fabric manipulation has generally focused on designing policies that are only applicable to a \textit{specific} task via manual design \cite{laundry2012,folding_iros_2015,folding_2017,shibata2012trajectory} or policy learning~\cite{seita_ryan, lerrel}. 

The difficulty in developing accurate analytical models of highly deformable objects such as fabric motivates using data-driven strategies to estimate models, which can then be used for general purpose planning. While there has been prior work in system identification for robotic manipulation~\cite{GP-MPC,cautious-MPC,berkenkamp2016safe, MPCRacing, system-id, handful-of-trials}, many of these techniques depend on reliable state estimation from observations, which is especially challenging for deformable objects. One recent alternative to system identification is visual foresight~\cite{visual_foresight_2018,finn_vf_2017}, which uses a large amount of self-supervised interactions to learn a visual dynamics model directly from raw image observations and has shown the ability to generalize to a wide variety of conditions~\cite{robonet}. This learned model can then be used for planning to perform different tasks at test time. The technique has been successfully applied to learning the dynamics of complex tasks such as pushing rigid objects~\cite{finn_vf_2017} and basic fabric folding~\cite{visual_foresight_2018}. However, two limitations of prior work in visual foresight are 1) the data requirement for learning accurate visual dynamics models is often very high, requiring several days of continuous data collection on real robots~\cite{robonet,visual_foresight_2018}, and 2) experiments consider only relatively short horizon tasks with a wide range of valid goal images~\cite{visual_foresight_2018}.

\begin{figure}[t]
\center
\includegraphics[width=0.75\textwidth]{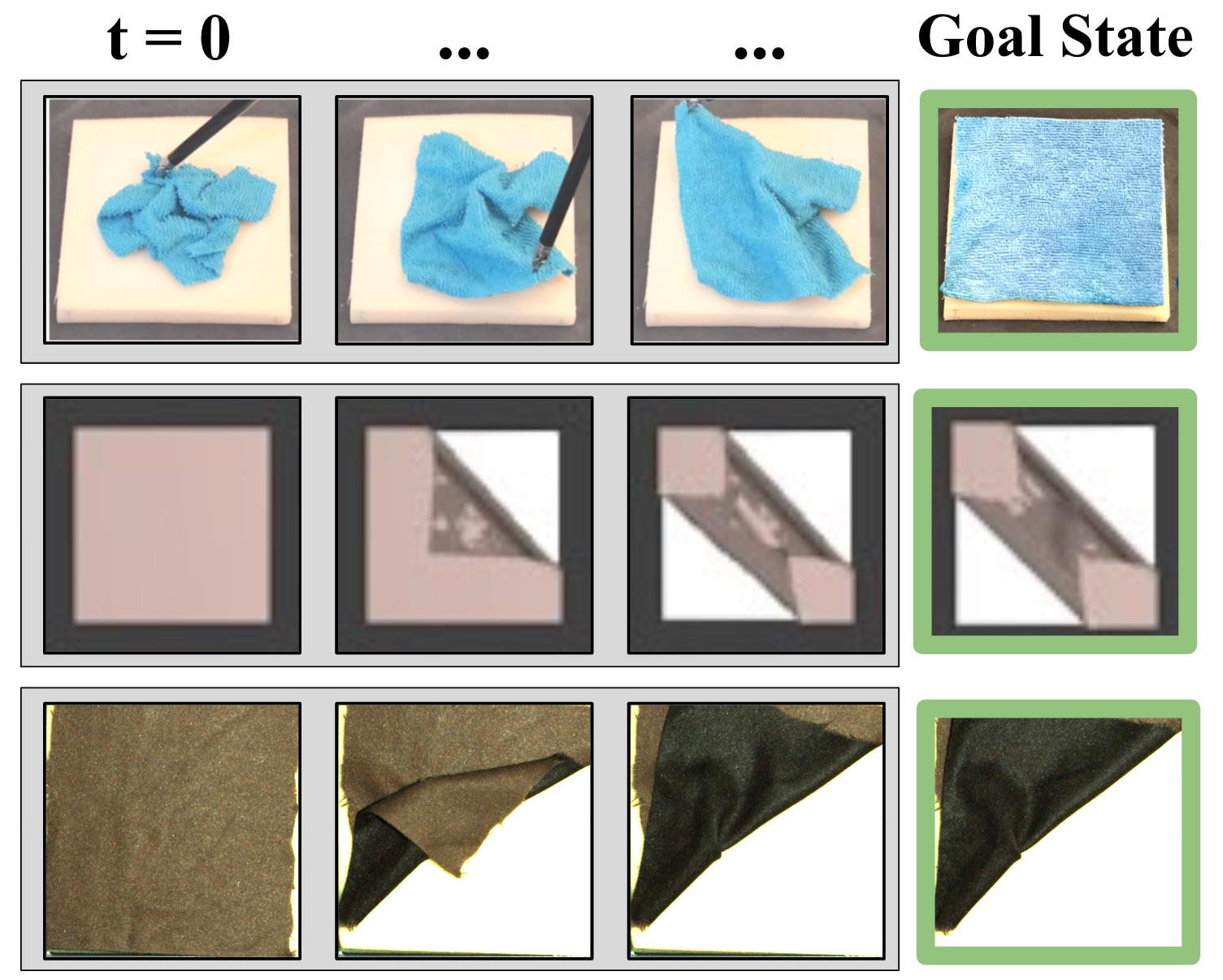}
\caption{
Using VSF trained on simulated RGBD data, we learn a goal-conditioned fabric manipulation policy without any task demonstrations. \textbf{Top:} Subsampled frames from a smoothing episode on physical fabric with the da Vinci surgical robot. \textbf{Middle:} Subsampled frames from a double folding episode in simulation with a new dataset, new optimizer, and new cost function. \textbf{Bottom:} Subsampled frames from a new folding episode on physical fabric.
}
\vspace*{-10pt}
\label{fig:teaser}
\end{figure}
 
This paper is an extended version of our prior work,~\citet{vsf-fabric}, which presented \vsf (VSF) by integrating RGB and depth sensing to learn and plan over dynamics models in simulation using only random interaction data for training and domain randomization techniques for sim-to-real transfer. That paper applied VSF to smoothing and folding tasks (see Figure~\ref{fig:teaser} for example rollouts). In this work, we explore modifications to all major stages of \vsf: the data generation, visual dynamics, cost function, and optimization procedure. Specifically, we make the following extensions:
\begin{enumerate}
\item A new dataset of 9,932 episodes for learning visual dynamics models for fabric manipulation with a corner selection bias and increased range of motion.
\item New simulation experiments evaluating the tradeoffs between different datasets, learned dynamics models, cost functions, and optimization procedures on system performance.
\item New physical experiments demonstrating 90\% reliability on fabric folding with a da Vinci surgical robot, a task the robot was unable to perform successfully in the prior work~\cite{vsf-fabric}.
\end{enumerate}
Results suggest that the most beneficial extension is the new dataset containing actions that have longer pull distances and bias towards picking at corners. This leads to larger changes in the fabric configuration in regions more broadly relevant for manipulation, and the learned dynamics models enable more reliable and efficient fabric folding on the physical robotic system.

%% file: 2-related-work.tex
\section{Related Work}
\label{sec:rw}

\subsection{\textbf{Geometric Approaches for Robotic Fabric Manipulation}}
Manipulating fabric is a long-standing challenge in robotics. In particular, prior work has focused on fabric smoothing, as it helps standardize the configuration of the fabric for subsequent tasks such as folding~\cite{grasp_centered_survey_2019,manip_deformable_survey_2018}. One popular approach in these works is to first hang fabric in the air and allow gravity to ``vertically smooth'' it~\cite{osawa_2007,kita_2009_iros,kita_2009_icra,unfolding_rf_2014}.~\citet{maitin2010cloth}, use this approach to achieve a 100\% success rate in single-towel folding over 50 trials. For tasks involving larger fabrics like blankets~\cite{seita-bedmaking}, or those utilizing single-armed robots with a limited range of motion, such vertical smoothing may be infeasible. An alternative approach is to perform fabric smoothing on a flat surface using sequential planar actions as in~\cite{heuristic_wrinkles_2014, cloth_icra_2015, willimon_unfolding_laundry_2011}, but these works assume initial fabric configurations closer to fully smoothed than those considered in this work. Similar work addresses both fabric smoothing and folding, such as by~\citet{balaguer2011combining} and Jia~et~al.~\cite{jia_visual_feedback_2018,jia_cloth_manip_2019}. These works assume the robot is initialized with the fabric already grasped, while we initialize the robot's end-effector away from the fabric.

\subsection{\textbf{Learning Fabric Manipulation in Simulation and in Real}}
There has been recent interest in learning sequential fabric manipulation policies with fabric simulators. For example,~\citet{seita_ryan} and~\citet{lerrel} learn fabric smoothing in simulation, the former using DAgger~\cite{ross2011reduction} and the latter using model-free reinforcement learning (MFRL). Similarly,~\citet{sim2real_deform_2018} and~\citet{rishabh_2019} learn policies for folding fabrics using MFRL augmented with task-specific demonstrations. All of these works obtain large training datasets from fabric simulators; examples of simulators with support for fabric include ARCSim~\cite{arcsim2012}, MuJoCo~\cite{mujoco}, PyBullet~\cite{coumans2019}, Blender~\cite{blender}, and NVIDIA FLeX~\cite{lin2020softgym}. While these algorithms achieve impressive results, they are designed or trained for specific fabric manipulation tasks such as folding or smoothing, and do not reuse learned structure to generalize to a wide range of tasks. 

\citet{ganapathi2020learning, mmgsd} generalize to multiple tasks by learning fabric correspondences in simulation, but require a task demonstration at test time. Other recent work such as~\cite{seita2020learning},~\cite{lin2020softgym} and~\cite{erickson2020assistivegym} also aim to learn generalizable fabric manipulation policies in simulation but focus more on rearrangement, transportation and dressing tasks respectively rather than complex folding tasks. \citet{lee2020learning} successfully learn an arbitrary goal-conditioned fabric folding policy in a model-free manner that is able to achieve unseen fabric goal configurations at test time, but the policy is learned entirely on a real robot which may cause wear-and-tear on the physical system.

\subsubsection{\textbf{Model-Based Fabric Manipulation}}
Combining model-predictive control (MPC) with learned dynamics is a popular approach for robotics control that has shown success in learning robust closed-loop policies even with substantial dynamical uncertainty~\cite{converging-supervisor,deep_dressing_2018,rosen_icra_tissues_2019,thananjeyan2019safety,recovery-rl}. However, many of these prior works require knowledge or estimation of underlying system state, which can often be inaccurate, especially for highly deformable objects. As an alternative to estimating the system state,~\citet{finn_vf_2017} and~\citet{visual_foresight_2018} introduce \textit{visual foresight}, and demonstrate how MPC can plan over learned video prediction models to accomplish a variety of robotic tasks, including deformable object manipulation such as folding pants. However, the trajectories shown in \citet{visual_foresight_2018} are limited to a single pick and pull, while we focus on longer horizon sequential tasks that are enabled by a pick-and-pull action space rather than direct end effector control. Furthermore, the fabric manipulation tasks reported have a wide range of valid goal configurations, such as covering a utensil with a towel or moving a pant leg upwards. In contrast, we focus on achieving precise goal configurations via multi-step interaction with the fabric. \citet{yan2020learning} also take a model-based approach to fabric manipulation, and primarily focus on the fabric smoothing task using latent dynamics models.~\citet{latentroadmap} generate action plans in a low-dimensional latent space and evaluate on a single T-shirt folding task.

This paper is a direct extension of Hoque~et~al.~\cite{vsf-fabric}, which learns a fabric dynamics model from RGB and depth images entirely in simulation and performs model-based planning over the learned dynamics model to achieve fabric smoothing and limited fabric folding tasks. This paper modifies and ablates various components of the pipeline in~\cite{vsf-fabric} including the data generation process, video prediction model, cost function, and action sampling method to improve the folding performance in both simulation and on a physical robotic system. 
 
\subsubsection{\textbf{Planning with Visual Dynamics Models}}
Prior work on visual foresight~\cite{finn_vf_2017, visual_foresight_2018, robonet} generally collects data for training visual dynamics models in the real world, which is impractical and unsafe for robots such as the da Vinci surgical robot due to the sheer volume of data required for the technique (on the order of 100,000 to 1 million actions, often requiring several days of physical interaction~\cite{robonet}). One recent exception is the work of~\citet{time_reversal}, which trains visual dynamics models in simulation for Tetris block matching. Finally, prior work in visual foresight learns visual dynamics models with RGB images, but we find that training with RBGD images improves performance, as depth data can provide valuable geometric information for fabric manipulation tasks involving multiple layers. 

%% file: 3-prob-statement.tex
\section{Problem Statement}
\label{sec:bg}
We consider learning goal-conditioned fabric manipulation policies that enable planning to specific fabric configurations given a goal image of the fabric in the desired configuration. The fabric lies on top of a flat background plane. We assume that the fabric shape is square and that the sides of the fabric are colored differently, where each side is monochromatic. In this paper we test on three goals: a smooth configuration, a triangular single-folded configuration, and a double-folded configuration with three layers stacked in the center of the image, as shown in Figure~\ref{fig:teaser}.

We define the fabric configuration at time $t$ as $\boldsymbol{\xi}_t$, represented via a mass-spring system with an $N \times N$ grid of point masses subject to gravity and Hookean spring forces. Due to the difficulties of state estimation for highly deformable objects, we consider overhead RGBD observations $\bo_t \in \mathbb{R}^{56 \times 56 \times 4}$, which consist of three-channel RGB and single-channel depth images.

Each task is specified with a goal image observation $\bo^{(g)} \in \mathbb{R}^{56 \times 56 \times 4}$ representing the goal $g$ which indicates the appearance of the world the robot must achieve after interacting with the fabric within some finite time horizon $T$. Thus, we only consider tasks which can be defined with an image of the goal configuration of the fabric. We further assume that the tasks can be achieved with a sequence of pick-and-place actions with a single robot arm, which involve grasping a specific point on the top layer of the fabric and pulling it in a particular direction. The above assumptions hold for a variety of common manipulation tasks such as folding and smoothing. We consider four-dimensional actions, 
\begin{equation}\label{eq:action}
\ba_t = \langle x_t, y_t, \Delta x_t, \Delta y_t\rangle.
\end{equation}
Each action $\ba_t$ at time $t$ involves grasping the top layer of the fabric at coordinate $(x_t, y_t)$ with respect to an underlying background plane, lifting, translating by $(\Delta x_t, \Delta y_t)$ while keeping height fixed, and then releasing and letting the fabric settle. 
When appropriate, we omit the time subscript $t$ for brevity.

The objective is to learn a goal-conditioned policy which minimizes some goal-conditioned cost function $c_g(\tau$) defined on realized interaction episodes with goal $g$ and episode $\tau = (\bo_1, \ldots, \bo_T)$, consisting of a sequence of image observations of the fabric.

%% file: 4-approach.tex
\section{\vsf}
\label{sec:approach}

We build on the visual foresight framework introduced by~\citet{finn_vf_2017} to learn goal-conditioned fabric manipulation policies. In visual foresight, a video prediction model (also called a visual dynamics model) is trained on random interaction data of the robot in the environment. This model is trained to generate a sequence of predicted images (i.e., frames) that would result from executing a sequence of proposed actions in the environment given a history of observed images. Then, MPC is used to plan over this visual dynamics model with some cost function evaluating the discrepancy between predicted images and a desired goal image. 

In~\citet{vsf-fabric}, we present \vsf, where 1) a visual dynamics model is trained on RGBD images instead of RGB images as in~\cite{finn_vf_2017}, and 2) visual dynamics are learned entirely in simulation. We find that these choices improve performance on complex fabric manipulation tasks in simulation and real, accelerate data collection, and limit wear-and-tear on a physical robot. In this work, we extend~\citet{vsf-fabric} and explore the tradeoffs involved in each of several different design decisions for each core aspect of \vsf. As elaborated later in Section~\ref{sec:results}, we use the term \vsfrss to refer to the specific settings used in~\cite{vsf-fabric}. In this section, we review learning VisuoSpatial dynamics models (Section~\ref{ssec:dynamics}), model-based planning over the learned dynamics model (Section~\ref{ssec:planning}), and specifying planning costs (Section~\ref{ssec:costs}). Each subsection discusses the methodology from our prior work~\cite{vsf-fabric} and then new alternative techniques that we explore in this paper. 

\subsection{Learning VisuoSpatial Dynamics}
\label{ssec:dynamics}

To represent fabric dynamics, we train deep recurrent convolutional networks~\cite{Goodfellow-et-al-2016} to predict a sequence of RGBD output images conditioned on a sequence of RGBD context images and a sequence of actions. As noted in~\citet{sv2p}, video prediction is inherently stochastic due to incomplete information provided from context images. For example, a pick-and-pull action applied on fabric will have different effects based on unknown stiffness and friction parameters. Therefore, we leverage two widely-used recurrent stochastic video prediction models: Stochastic Variational Video Prediction (SV2P) from~\citet{sv2p}, which we used in~\cite{vsf-fabric}, and Stochastic Video Generation (SVG) from~\citet{svg}, a more recent model which we evaluate in this work. We describe details in Sections~\ref{sssec:sv2p} and~\ref{sssec:svg}, respectively.

\subsubsection{Stochastic Variational Video Prediction}\label{sssec:sv2p}

SV2P is an action-conditioned video prediction model which can predict future images conditioned on a sequence of prior images and proposed actions. In~\cite{vsf-fabric}, we trained SV2P in a self-supervised manner on thousands of episodes of random interaction with the fabric from the simulation environment in~\citet{seita_ryan}, where an episode consists of a contiguous trajectory of observations and pick-and-pull actions (Equation~\ref{eq:action}).

Precisely, SV2P trains a generative model to predict a sequence of $H$ output images conditioned on a context vector of $m$ images and a sequence of actions starting from the most recent context image. To handle stochasticity, SV2P uses latent variables to capture different modes in the distribution of predicted images, thus making predictions conditioned on a vector of latent variables $\bz_{t+m:t+m+H-1}$, each sampled from a Gaussian prior distribution at inference time. For SV2P, the prior $p(\bz)$ is fixed at each time step $t$, resulting in the generative model parameterized by $\theta$:
\begin{align}
\begin{split}\label{eq:sv2p}
    &p_\theta(\hat{\bo}_{t+m:t+m+H-1} | \hat{\ba}_{t+m-1:t+m+H-2}, \bo_{t:t+m-1}) \\ 
    &\hspace{5mm} = p(\bz_{t+m}) \prod_{t'= t + m}^{t + m + H - 1} p_\theta(\hat{\bo}_{t'} | \bo_{t:t+m-1}, \hat{\bo}_{t+m:t' - 1}, \bz_{t'}, \hat{\ba}_{t'-1}).
\end{split}
\end{align}
Here $\bo_{t:t+m-1}$ are image observations from time $t$ up to and including $t+m-1$, $\hat{\ba}_{t+m-1:t+m+H-2}$ is a candidate action sequence at timestep $t+m-1$, and $\hat{\bo}_{t+m:t+m+H-1}$ is the sequence of predicted images. Since the generative model is trained in a recurrent fashion, it can be used to sample an $H$-length sequence of predicted images $\hat{\bo}_{t+m:t+m+H-1}$ for any $m > 0, H > 0$ conditioned on a current sequence of image observations $\bo_{t:t+m-1}$ and an $H$-length sequence of proposed actions taken from $\bo_{t+m-1}$, given by $\hat{\ba}_{t+m-1:t+m+H-2}$. For more details on model architectures and training procedures for SV2P, we refer the reader to~\citet{sv2p}. We build upon author-provided open-source code for SV2P from~\cite{tensor2tensor}.

\subsubsection{Stochastic Video Generation}\label{sssec:svg}

\begin{figure}[t]
\center
\includegraphics[width=1.00\textwidth]{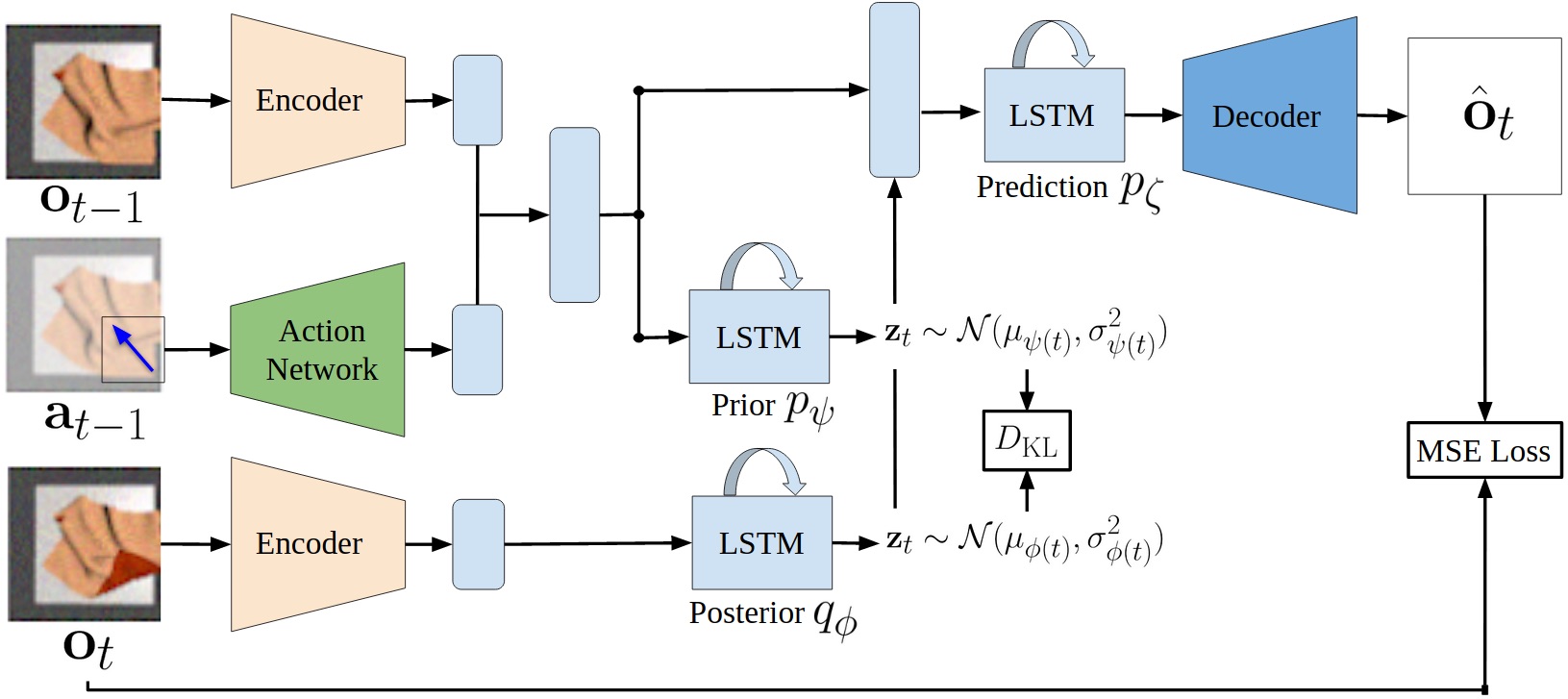}
\caption{
Flow of data through the proposed action-conditioned SVG architecture during training, described in Section~\ref{sssec:svg}. Given the most recent context image $\bo_{t-1}$ and the action $\ba_{t-1}$ taken at that time (visualized with the overlaid arrow), the model is trained to predict the next image $\bo_t$ via a Mean Square Error loss while simultaneously minimizing a KL divergence loss between the prior $p_\psi$ and posterior $q_\phi$ Gaussian distributions. The encoder and decoder are convolutional networks with architectures similar to DCGAN. The prior $p_\psi$, posterior $q_\phi$, and prediction $p_\zeta$ networks use LSTMs. Each action $\ba_{t-1} \in \mathbb{R}^4$ is passed through a small learned network, whose output is concatenated with the encoder output of $\bo_{t-1}$. During \emph{training}, the latent variable $\bz_t$ is sampled from the output of the posterior distribution, but during \emph{inference} time, the posterior is removed and $\bz_t$ is sampled from the prior. 
}
\vspace*{-10pt}
\label{fig:svg-action}
\end{figure}

As an alternative to SV2P, we test with SVG~\cite{svg}, which has been found to predict sharper images on standardized benchmarks such as Stochastic Moving MNIST~\cite{svg} and the BAIR Robot Pushing dataset~\cite{bair_push_2017}. SVG, however, does not support action conditioning, which is critical for model-based planning. We add support for action conditioning to SVG, as described below, with a similar approach as in prior work from~\citet{gap_suraj_2020}.
During training, similarly to SV2P, SVG samples latent variables $\bz_t$ from a Gaussian posterior distribution; however, while SV2P samples $\bz_t$ from the \emph{same} distribution for each $t$, SVG samples from a \emph{different}, time-dependent posterior distribution $q_\phi(\bz_t | \bo_{1:t})$ with parameters $\phi$ (where the dependence on $t$ makes it time-dependent). At inference time, SVG samples $\bz_t$ from a Gaussian prior distribution, similarly to SV2P. Unlike SV2P's approach of sampling from a fixed prior $p(\bz_t)$ for each $t$ (see Equation~\ref{eq:sv2p}), SVG uses a more flexible, time-varying prior distribution $p_\psi(\bz_t | \bo_{1:t-1})$ with parameters $\psi$ learned during training. \citet{svg} argue that these more flexible distributions lead to better video prediction quality. We refer the reader to~\cite{svg} for further details. 

Figure~\ref{fig:svg-action} shows the flow of image and action data through the proposed action-conditioned SVG variant used in this work. We use DCGAN-style~\cite{dcgan2016} encoders and decoders to handle the image embedding, along with generic LSTMs~\cite{hochreiter1997long} for the prior $p_\psi$, posterior $q_\phi$, and frame predictor $p_\zeta$ components of SVG. To handle action conditioning, at each time step we feed actions (see Equation~\ref{eq:action}) through a small fully connected network with two layers, and then we concatenate the result with the embedded image from the encoder. The dimension of each embedded image is 128 and the output dimension of the action network is 32, resulting in a concatenated 160-dimensional vector, which is then propagated through the prior and prediction LSTMs. The action-conditioned SVG has a total of 12.6 million parameters, as compared to about 7.9 million parameters for SV2P.

\subsection{Model-Based Planning with Model Predictive Control}
\label{ssec:planning}
The goal of the planning stage is to determine which action the robot should take at each time step $t$. At each step, \vsf minimizes a goal-conditioned planning cost function $c_g(\hat{\bo}_{t+1:t+H})$ with goal $g$, which is a target image $\bo^{(g)}$. The cost is evaluated over the $H$-length sequence of predicted images $\hat{\bo}_{t+1:t+H}$ sampled from the visuospatial dynamics model (see Section~\ref{ssec:dynamics}) conditioned on the current observation $\bo_t$ and some proposed action sequence $\hat{\ba}_{t:t+H-1}$. As in prior Visual Foresight work~\cite{robonet,visual_foresight_2018,finn_vf_2017}, we utilize MPC to plan action sequences to minimize $c_g(\hat{\bo}_{t+1:t+H})$ over a receding $H$-step horizon at each time $t$. See Figure~\ref{fig:planning} for intuition on the planning phase in the context of fabric manipulation using \vsfrss from prior work~\cite{vsf-fabric}.

There are a number of sampling-based, gradient-free methods to optimize the MPC objective. In our prior work~\cite{vsf-fabric}, we used the Cross Entropy Method (CEM)~\cite{cem_1999}, which we review in Section~\ref{sssec:cem}. In this work, we additionally test with the Covariance Matrix Adaptation Evolution Strategy (CMA-ES)~\cite{CMA-ES} (see Section~\ref{sssec:cma}), which allows for more rapid updates in the action sampling distribution. While the Model-Predictive Path Integral (MPPI) has also been shown to be successful in recent work~\cite{nagabandi_2018}, we find that its temporal smoothing is not well-suited for our action space, in which we specify large pick-and-pull actions that are not expected to be similar to prior actions in a given trajectory.

\subsubsection{Cross Entropy Method (CEM)}\label{sssec:cem}
In a high-dimensional space, sampling actions uniformly at random is unlikely to yield a high-quality solution to an optimization problem. To mitigate this issue, we use CEM, which samples from a multivariate Gaussian distribution and iteratively re-fits the Gaussian to the best performing samples. Specifically, for each iterations, CEM:
\begin{enumerate} 
\item Samples $N$ action sequences $\{\hat{\ba}^{(1)}, \hat{\ba}^{(2)}, \dots , \hat{\ba}^{(N)}\}$ from some $\mathcal{N}(\mu_i, \Sigma_i)$
\item Finds the $M$ best sequences $\{\hat{\ba}^{(1)}, \hat{\ba}^{(2)}, \dots , \hat{\ba}^{(M)}\}$ according to cost $c_g(\cdot)$
\item Assigns $\mu_{i+1} = $ \textrm{mean}($\{\hat{\ba}^{(1)}, \hat{\ba}^{(2)}, \dots , \hat{\ba}^{(M)}\}$)
\item Assigns $\Sigma_{i+1} = $ \textrm{var}($\{\hat{\ba}^{(1)}, \hat{\ba}^{(2)}, \dots , \hat{\ba}^{(M)}\}$)
\end{enumerate}
where ``mean$(\cdot)$'' and ``var$(\cdot)$'' denote the sample mean and covariance. However, CEM still scales poorly with dimensionality and can struggle with multimodal optimization landscapes due to its Gaussian structure.

\subsubsection{Covariance Matrix Adaptation Evolution Strategy (CMA-ES)}\label{sssec:cma}
Like CEM, CMA-ES is an evolutionary strategy based on sampling actions from an iteratively re-fitted multivariate Gaussian distribution. However, in CEM, the standard deviations of the sampling distribution in consecutive iterations are highly correlated, making it difficult to rapidly adjust the variability of the sampling distribution. CMA-ES mitigates this issue by fitting a full covariance matrix to the elite samples and using it to tune the variance of the sampling distribution on each iteration in a more fine-grained manner. CMA-ES also updates the mean to a weighted average over the elites rather than a simple average. Additionally, while CEM updates a \textit{large} population over a \textit{small} number of iterations, we run CMA-ES with a \textit{small} population over a \textit{large} number of iterations, resulting in a more dynamic search of the optimization landscape less prone to averaging over multiple modes. We refer the reader to~\cite{CMA-ES} for further details and to Appendix~\ref{app:visualmpc} for exact parameters used for both CEM and CMA-ES. 
\begin{figure}[t]
\center
\includegraphics[width=0.90\textwidth]{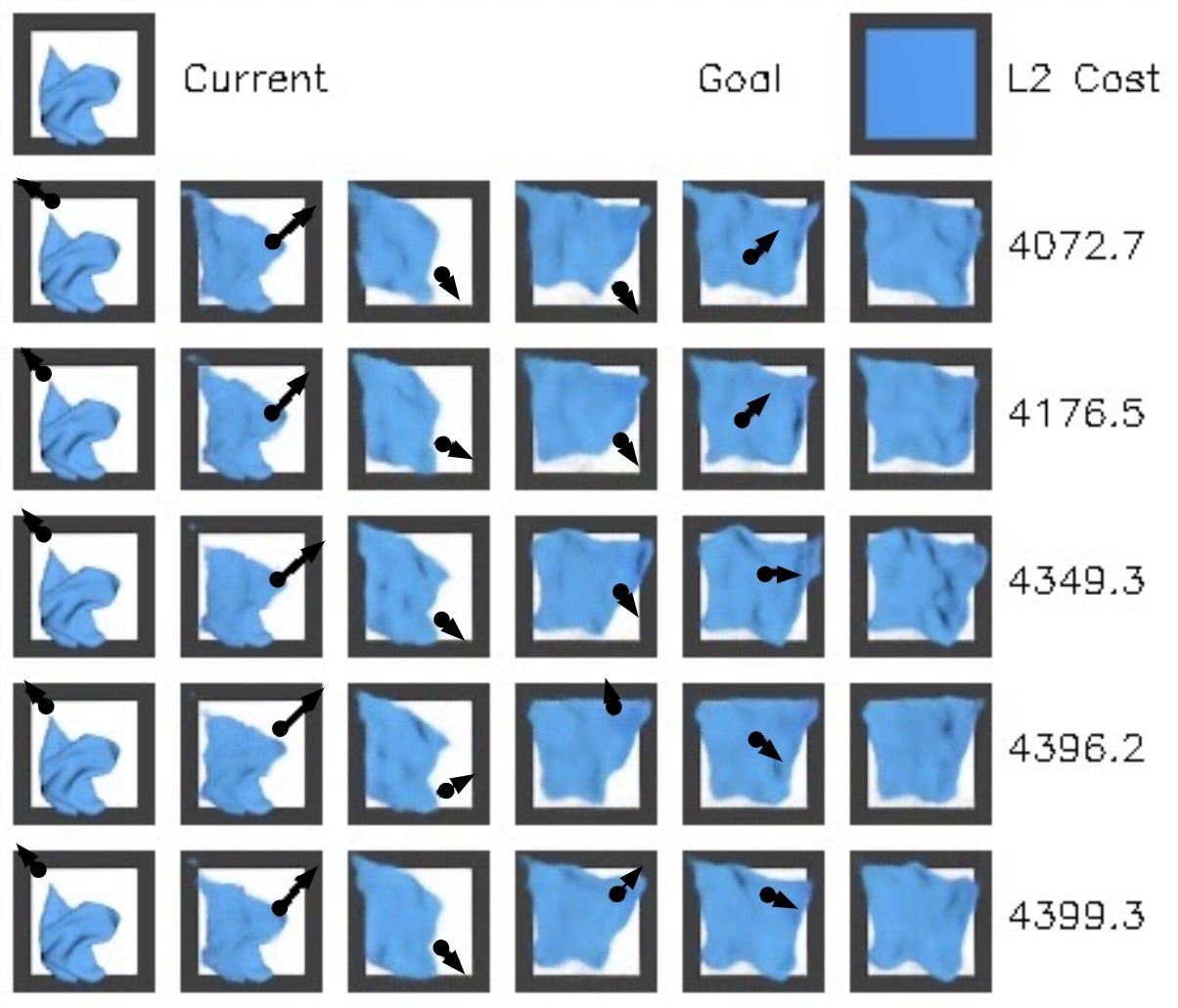}
\caption{
Real plans generated by \vsfrss at test time for the smoothing task using \dold (see Section~\ref{ssec:data-gen}) with predictions from SV2P. We generate action sequences with the Cross Entropy Method (CEM) to approximately minimize the cost function, which evaluates L2 distance between the final image in the predicted trajectory and a provided goal image (see Section~\ref{sssec:l2-cost}). Here we show the five CEM trajectories with the lowest cost, where the image sequences in each row first show the context image, followed by five outputs from the video prediction model. The black arrows are the pick-and-pull actions projected onto the images.
}
\vspace*{-10pt}
\label{fig:planning}
\end{figure}

\subsection{Planning Costs}
\label{ssec:costs}

The remaining ingredient for model-based planning (Section~\ref{ssec:planning}) is the cost function definition. A number of choices exist for the cost function $c_g(\cdot)$ used for model-based planning based on goal classifiers, optical flow, and learned distance measures between images~\cite{visual_foresight_2018, fewshot_xie_2018}. In our prior work~\cite{vsf-fabric}, we utilized a simple cost function based on the Euclidean pixel distance between images (Section~\ref{sssec:l2-cost}). In this work, we additionally test with a learned cost function which encodes structure about the underlying mesh of the fabric (Section~\ref{sssec:cost2}).

\subsubsection{Pixel L2 Cost}\label{sssec:l2-cost}
A simple cost function is the Euclidean (L2) pixel distance between the final predicted RGBD image at timestep $t$ and the goal image $\bo^{(g)}$ across all 4 channels. Precisely, the planning cost is defined as follows:
\begin{equation}\label{eq:cost}
    c_g(\hat{\bo}_{t+1:t+H}) = \|\bo^{(g)} - \hat{\bo}_{t+H}\|_2.
\end{equation}
Figure~\ref{fig:planning} shows example plans generated by the system for smoothing and indicates the L2 cost between the final predicted image and the goal image in the rightmost column. We use ``Pixel L2'' to refer to this cost function in this paper.

\subsubsection{Learned Vertex L2 Cost}\label{sssec:cost2}
While the Pixel L2 distance cost function is easy to implement and may be sufficient for simple goal images such as fully smooth fabric, it can fail to capture nuances and may focus on irrelevant artifacts in more complex goal images. To this end, we employ a data-driven approach to estimate the cost between two images. When collecting data used to train the visuospatial dynamics model, we can access and store the underlying state of the fabric due to the simulation environment. Therefore, we utilize the same data to train a cost function which estimates the difference in the underlying fabric state based on two images of the fabric in different configurations. Precisely, we annotate pairs of fabric images with the total Euclidean distance between the 3D meshes that constitute the fabric (see Section~\ref{ssec:sim}) in each image. We then train a convolutional neural network $f_{\rm mesh}(\cdot, \cdot)$ to predict the (normalized) mesh distance from images by minimizing the Mean Squared Error (MSE) loss on the dataset. The revised planning cost function takes a forward pass through this trained network:
\begin{equation}\label{eq:cost2}
    c_g(\hat{\bo}_{t+1:t+H}) = f_{\rm mesh}(\bo^{(g)}, \hat{\bo}_{t+H}),
\end{equation}
and since \vsf data is task-agnostic, as described in Section~\ref{ssec:data-gen}, we use the same $f_{\rm mesh}$ network for all three major tasks considered in this work: smoothing, single folding, and double folding. We use ``Vertex L2'' to refer to this cost function. See Appendix~\ref{app:visualmpc} for details on the supplemental dataset and network architecture used for $f_{\rm mesh}$.

%% file: 4.5-implementation.tex
\section{Practical Implementation Details}
\label{sec:implementation}

In this section, we provide additional details to practically instantiate \vsf for goal-conditioned fabric manipulation. We discuss the fabric simulator used for data collection (Section~\ref{ssec:sim}), how data is collected in this simulator (Section~\ref{ssec:data-gen}), and how to train the visuospatial dynamics models (Section~\ref{ssec:modeltraining}).

\subsection{Fabric Simulator}
\label{ssec:sim}

\vsf requires a large amount of training data to predict full-resolution RGBD images. Since getting real data is cumbersome and imprecise, we use a fabric simulator to generate data quickly and efficiently. The fabric and robot simulator used in~\cite{vsf-fabric} and this work is built on top of the simulator in~\citet{seita_ryan}, which was shown to be sufficiently accurate for imitation learning and sim-to-real transfer of fabric smoothing policies. We briefly review details of the simulator that are shared across both~\cite{vsf-fabric} and this work, while highlighting differences in Section~\ref{ssec:data-gen}.

The fabric is represented as a mass-spring system with a $25 \times 25$ grid of point masses~\cite{provot_1996} with springs connecting each point to its neighbors. Verlet integration~\cite{verlet_1967} is used to update point mass positions using finite difference approximations, and self-collision is implemented by adding a repulsive force between points that are too close~\cite{cloth-cloth-collisions}. Damping is also applied to simulate friction. See Appendix~\ref{app:simulators} for further discussion on alternative fabric simulators and the simulation used in this work.

We use the open-source software Blender~\cite{blender} to render (top-down) image observations $\bo_t$ of the fabric. To facilitate sim-to-real transfer, we leverage domain randomization~\cite{domain_randomization} of the fabric color, background plane shading, image brightness, and camera pose. We make a few changes to the observations relative to~\cite{seita_ryan}. First, we use four-channel images: three for RGB and one for depth. Second, we reduce the size of observations to $56\times 56$ from $100\times 100$ to make it more computationally tractable to train visuospatial dynamics models. Finally, to enable transfer of policies from simulation to the real-world, we adjust the domain randomization techniques so that color, brightness, and positional hyperparameters are fixed \emph{per episode} to ensure that the video prediction model learns to only focus on predicting changes in the fabric configuration, rather than changes due to domain randomization. See Appendix~\ref{app:results} for more details on the domain randomization parameters. 

\subsection{Data Generation}
\label{ssec:data-gen}

For generating training data for \vsf, episode starting states are sampled from four difficulty tiers with equal probability, where each tier differs in the initial amount of fabric coverage on the underlying plane supporting it. Tiers 1 through 3 are the same as those in~\cite{seita_ryan}.
\begin{itemize}
\item \textbf{Tier 0: Full Coverage.} $100. \pm 0.$ initial coverage, i.e., fully smooth.
\item \textbf{Tier 1: High Coverage.} $78.3 \pm 6.9$\% initial coverage, all corners visible. Generated by two short random actions. 
\item \textbf{Tier 2: Medium Coverage.} $57.6 \pm 6.1$\% initial coverage, one corner occluded. Generated by a vertical drop followed by two actions to hide a corner.
\item \textbf{Tier 3: Low Coverage.} $41.1 \pm 3.4$\% initial coverage, 1-2 corners occluded. Generated by executing one action very high in the air and dropping. 
\end{itemize}

\subsubsection{\dold Data}
In the prior work~(\citet{vsf-fabric}), we collected a dataset consisting of 7,003 episodes of length 15 each, for a total of 105,045 ($\bo_t, \ba_t, \bo_{t+1}$) transitions. Actions $(x,y,\Delta x, \Delta y)$ are sampled uniformly from [0,0,0,0] to [1,1,0.4,0.4], allowing a maximum pull of 40\% of the plane width. Fabric color is randomized in a range around blue, and the underside of the fabric is darker by a fixed RGB delta. Henceforth, we refer to this data as \dold. Figures~\ref{fig:planning} and~\ref{fig:rollout_smooth} provide examples of images from \dold without domain randomization.

All data is generated using the following policy: execute a randomly sampled action, but resample if the grasp point $(x,y)$ is not within the bounding box of the 2D projection of the fabric, and truncate $\Delta x$ and/or $\Delta y$ at the edge of the plane if $(x + \Delta x, y + \Delta y)$ is out of bounds.

\subsubsection{New \dnew Data}\label{sssec:new-data-gen}

While we showed promising smoothing and folding simulation results using \dold in~\cite{vsf-fabric}, we were unable to get a physical robot to successfully fold fabric. In addition, the pull vector action magnitudes for \vsfrss were relatively small compared to an imitation learning baseline from~\cite{seita_ryan}. This meant \vsfrss was inefficient and took several more actions than necessary to complete smoothing or folding tasks. To address these issues, we propose and evaluate \dnew, a new fabric dataset with several notable differences over \dold.

\dnew consists of 9,932 length-10 episodes for a total of 99,320 ($\bo_t, \ba_t, \bo_{t+1}$) data transitions, meaning that the data is about the same size as \dold. For visual clarity, the fabric color in the new data is centered around brown (as opposed to blue in \dold). During data generation, actions $(x,y,\Delta x, \Delta y)$ are sampled from [0,0,0,0] to [1,1,0.6,0.6], allowing a maximum pull of 60\% of the plane width. While this increased range of motion may make subsequent video prediction more challenging, since longer pull vectors tend to result in larger relative pixel changes in future images, we hypothesize that including longer pull vectors in the training data will result in more accurate image predictions when considering such actions during MPC planning (see Section~\ref{ssec:planning}).

Many fabric manipulation tasks, including the smoothing and folding tasks we consider in this work, may be best approached by picking at fabric corners, as suggested by results in~\cite{lin2020softgym,seita-bedmaking,lerrel}. Therefore, we set 30\% of all pick points to be the $(x,y)$ coordinates of a randomly chosen corner, to which we have ground-truth access in the simulator. This ``corner bias'' is not present in \dold, which may have led \vsfrss to produce relatively less accurate future image predictions conditioned on actions that pick at corners. Due to this extra feature, we name the data ``\dnew.''

Dataset curation is an interesting topic in its own right. In general, the dataset should include states in regions that are relevant for the downstream tasks for reliable video prediction. It is difficult to reach states that require a precise sequence of actions with a purely random policy. In this case, the corner bias can help provide data broadly relevant for many smoothing and folding tasks. More complex tasks such as twisting or rolling fabric would require more careful dataset engineering.


Finally, to provide training data for the learned cost function (Section~\ref{sssec:cost2}) we also collect the ground truth $(x,y,z)$ coordinates of all 625 point masses for all time steps in all collected episodes. From \dnew we create a dataset of 99,320 RGBD image pairs annotated with ground truth mesh distance to train the learned cost function as described in Section~\ref{sssec:cost2}.

\subsection{Model Training}\label{ssec:modeltraining}

When training a visuospatial dynamics model (either SV2P or SVG) on \dold, as in~\cite{vsf-fabric} and following the notation from Equation~\ref{eq:sv2p}, we set the number of context frames to $m = 3$ and number of output frames to $H = 7$, so that the model learns to predict 7 frames of an episode from the preceding 3 frames. On \dnew, since the number of actions per episode in the training data is 10 (instead of 15), we set the number of context frames to $m = 2$ and the number of output frames to $H = 5$ to allow for sampling at multiple time ranges within one episode. At test time, both models utilize only one context frame $m = 1$ and a planning horizon of $H=5$ output frames. This yields the generative model $p_\theta(\hat{\bo}_{t+1:t+5} | \hat{\ba}_{t:t+4}, \bo_{t})$, as discussed in Section~\ref{ssec:dynamics}.

%% file: 5-simulated-exps.tex
\section{Simulation Experiments}\label{sec:results}
In this section, we report experimental results on the fabric simulation environment. In Section~\ref{ssec:pred}, we qualitatively and quantitatively analyze the performance of visuospatial dynamics models on predicting images in held-out test episodes, for all combinations of datasets (\dold and \dnew) and models (SV2P and SVG). Section~\ref{ssec:policy} presents results from our prior work~\cite{vsf-fabric} using \vsfrss settings: \dold data, SV2P, CEM, and Pixel L2 cost.  We then introduce new results in Section~\ref{ssec:auro-sim} to test whether changing any set of parameter settings from those in Section~\ref{ssec:policy} lead to better performance in downstream fabric manipulation tasks.
\begin{figure}[t]
\center
\includegraphics[width=0.9\textwidth]{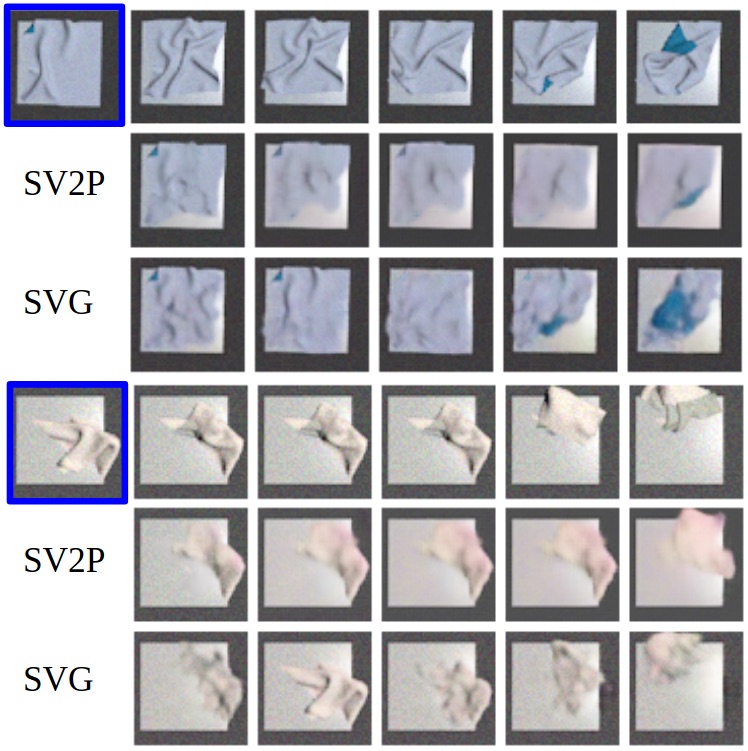}
\caption{
Two comparisons between ground truth and predicted color images in simulation from SV2P and SVG models on held-out, domain-randomized test data from \dnew. SV2P and SVG are provided a single context ground truth image (indicated with the blue border) and a sequence of 5 actions. For each example, the first row has the ground truth image sequence, the second shows SV2P predictions, and the third shows SVG predictions. While quality gets blurrier across time, the predicted images may be sufficiently accurate for planning. 
}
\vspace*{-10pt}
\label{fig:video-qualitative-color}
\end{figure}

\begin{figure}[t]
\center
\includegraphics[width=0.9\textwidth]{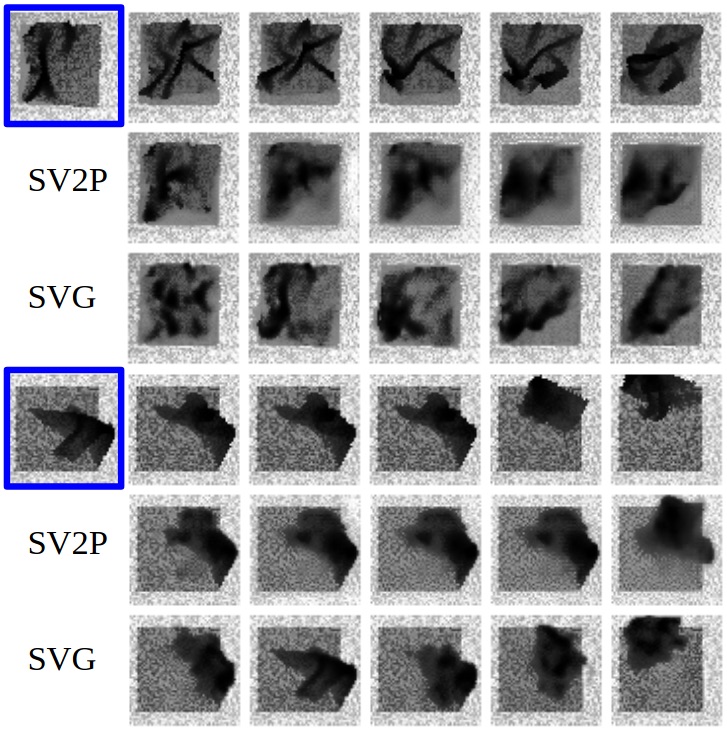}
\caption{
Depth components of the examples in Figure~\ref{fig:video-qualitative-color}, showing a similar trend. Depth values are scaled into $[0,255]$ to make images readable.
}
\vspace*{-10pt}
\label{fig:video-qualitative-depth}
\end{figure}

\subsection{VisuoSpatial Dynamics Prediction Quality}\label{ssec:pred}

An advantage of training visual dynamics models, as done in visual foresight methods, is that it enables inspection of models to see if predictions are accurate. We perform qualitative and quantitative analysis of action-conditioned video prediction model quality. For both \dold and \dnew data, we generate 400 episodes using the same data-generating procedure from Section~\ref{ssec:data-gen}, but with different random seeds to ensure that the test set contains novel images. We train separate SV2P and SVG models for both \dold and \dnew and evaluate each of these four models on the appropriate test set. 

The models are trained to directly generate RGBD predictions, and we separate the color and depth components for qualitative analysis. Figures~\ref{fig:video-qualitative-color} and~\ref{fig:video-qualitative-depth} show the ground truth as well as the predicted color and depth image sequences from SV2P and SVG, applied on examples of test-set episodes from \dnew. 
A prominent distinction between SV2P and SVG is that the former tends to produce blurrier images when predicting 4 or 5 images in the future as compared to SVG. However, SVG may be more susceptible to producing disjointed segments of the fabric. We hypothesize that this is because SV2P relies on an architecture which constrains the flow of predicted pixels~\cite{finn_cdna} while SVG does not.

For a more quantitative measure of prediction quality, we calculate the average Structural SIMilarity (SSIM) index~\cite{ssim_2004} over corresponding predicted and ground truth image pairs for the tested models. The SSIM is a scalar quantity between -1 and 1, where higher values correspond to greater image similarity. SSIM is commonly reported in prior video prediction research~\cite{sv2p,svg,finn_cdna,videoflow,savp} for quantitatively benchmarking model quality.
Tables~\ref{tab:video-quality-dold} and~\ref{tab:video-quality-dnew} report the performance of the two models on each of the two datasets. We report the average SSIM across predicted images as a function of the time horizon. As expected, SSIM decreases with a longer time horizon, due to the difficulty in long-horizon frame prediction. The results also suggest that SV2P tends to produce more accurate predictions for a shorter time horizon, typically the first 1-2 future images, while SVG may be more accurate for longer horizon predictions (4-5 images in the future).

Overall, the qualitative inspections and quantitative SSIM metrics suggest that using SV2P or SVG as the learned dynamics model may generate sufficiently accurate action-conditioned predictions for multiple images. 

\begin{table}[t]
\caption{
SSIM measurements for \dold, over ground truth versus predicted images from SV2P and SVG models. Conditioned on one image and five actions starting from that image, the models must predict the next five images. We separate SSIM measurements for color (C) and depth (D) images and by time horizon (i.e. 1-5 time steps into the future). Results suggest that SV2P is more effective at predicting the first 1-2 images, but SVG may produce more accurate predictions beyond that.
}
\centering
\begin{tabular}{ l r r r r r}
\textbf{(Data) Model} & 1 & 2 & 3 & 4 & 5\\ \hline 
(C) SV2P    & \textbf{0.822} & \textbf{0.710} & 0.638 & 0.611 & 0.598 \\
(C) SVG     & 0.755 & 0.682 & \textbf{0.648} & \textbf{0.633} & \textbf{0.624} \\ \hline
(D) SV2P    & \textbf{0.790} & \textbf{0.631} & 0.527 & 0.470 & 0.433 \\
(D) SVG     & 0.648 & 0.574 & \textbf{0.540} & \textbf{0.523} & \textbf{0.511} \\ \hline
\end{tabular}
\label{tab:video-quality-dold}
\end{table}

\begin{table}[t]
\caption{
SSIM measurements for \dnew, over ground truth versus predicted images from SV2P and SVG models. The table is formatted in a similar manner to Table~\ref{tab:video-quality-dold} and shows a similar trend.
}
\centering
\begin{tabular}{ l r r r r r}
\textbf{(Data) Model} & 1 & 2 & 3 & 4 & 5\\ \hline 
(C) SV2P & \textbf{0.774} & \textbf{0.706} & 0.639 & 0.616 & 0.605 \\
(C) SVG  & 0.741 & 0.667 & \textbf{0.642} & \textbf{0.631} & \textbf{0.625} \\ \hline
(D) SV2P & \textbf{0.758} & \textbf{0.657} & \textbf{0.577} & 0.529 & 0.493 \\
(D) SVG  & 0.679 & 0.602 & \textbf{0.577} & \textbf{0.563} & \textbf{0.554} \\ \hline
\end{tabular}
\label{tab:video-quality-dnew}
\end{table}

\subsection{Prior Results from Smoothing and Folding in Simulation}\label{ssec:policy}
All results presented in this section are from our prior work~\cite{vsf-fabric}. We report the performance of VSF with the \dold dataset, SV2P model, CEM optimizer, and Pixel L2 cost function. We refer to these particular choices of the data, model, optimizer, and cost function of VSF as \vsfrss, to distinguish these settings from different ablations we test in new experiments in Section~\ref{ssec:auro-sim}. We first evaluate \vsfrss on the smoothing task: maximizing fabric coverage, defined as the percentage of an underlying plane covered by the fabric. The plane is the same area as the fully smoothed fabric. We evaluate smoothing on three tiers of difficulty as reviewed in Section~\ref{ssec:data-gen} (i.e., tiers 1, 2, and 3). Following our prior work~\cite{vsf-fabric}, episodes can terminate earlier if a threshold of 92\% coverage is triggered, or if any fabric point falls sufficiently outside of the fabric plane.

\begin{figure*}[t]
\center
\includegraphics[width=1.00\textwidth]{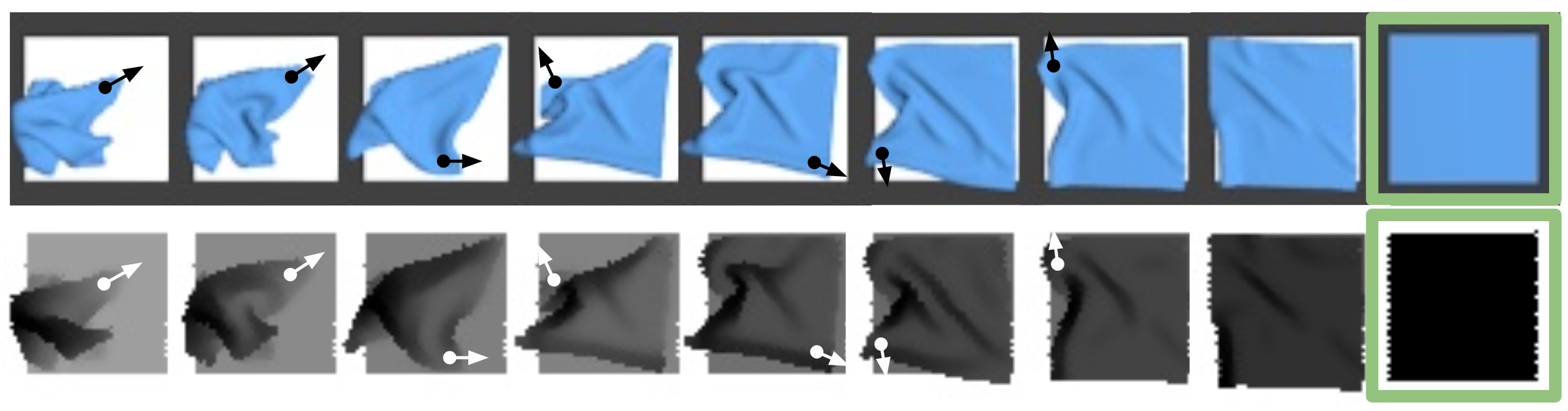}
\caption{
A simulated episode executed by the \vsfrss policy on a Tier 3 starting state, given a smooth goal image (shown in the far right). The first row shows RGB images and the second shows the corresponding depth maps. The images are from the distribution specified in the \dold data and do not have domain randomization. In this example, the policy is able to successfully cross the coverage threshold of 92\% after executing 7 actions. Actions are visualized with the overlaid arrows.
}
\vspace*{-10pt}
\label{fig:rollout_smooth}
\end{figure*}

To see how \vsf performs against existing smoothing techniques, for each difficulty tier, we execute 200 episodes of \vsfrss and 200 episodes of each baseline policy discussed in Section~\ref{sssec:baseline-methods}. Note that \vsfrss does \textit{not} explicitly optimize for coverage and only optimizes the Pixel L2 cost function from Equation~\ref{eq:cost}, which measures Euclidean distance to a target image. In this case, we provide \vsfrss with a goal image of a fully smooth fabric. See Figure~\ref{fig:rollout_smooth} for an example smoothing episode.

\subsubsection{Baseline Methods}\label{sssec:baseline-methods}

For fabric smoothing in simulation, we compare \vsfrss with the following 5 baselines as in~\citet{vsf-fabric}.
Further details about the implementation and training of \vsfrss and the last two baselines listed here are in Appendix~\ref{app:implementation}.

\paragraph{(1) Random} Randomly sample the pick point and pull direction.
\paragraph{(2) Highest} Using ground truth state information, pick the fabric vertex with the maximum $z$-coordinate and set the pull direction to point to where the vertex would be if the fabric were perfectly smooth. This is straightforward to implement with depth sensing and was shown to work reasonably well for smoothing in~\cite{seita-bedmaking}.
\paragraph{(3) Wrinkle} As in~\citet{cloth_icra_2015}, find the largest wrinkle and then pull perpendicular to it at the edge of the fabric to smooth it out. We use the ground truth state information in the implementation of this algorithm (as done in~\cite{seita_ryan}) rather than image observations.
\paragraph{(4) Imitation Learning (IL)} As in~\citet{seita_ryan}, train an imitation learning agent using DAgger~\cite{ross2011reduction} with a simulated corner-pulling demonstrator that picks and pulls at the fabric corner furthest from its target. DAgger can be considered as an oracle with ``privileged'' information as in~\citet{cheating_2019} because during training, it queries a demonstrator which uses ground truth state information. For a fair comparison, we run DAgger so that it consumes roughly the same number of data points (we used 110,000) as \vsf during training, and we give the policy access to four-channel RGBD images. We emphasize that this is a distinct dataset from the one used for \vsfrss or any other \vsf variant in this subsection (\dold), which uses no demonstrations during data generation.
\paragraph{(5) Model-Free RL} We run DDPG~\cite{ddpg2016} and extend it to use demonstrations and a pre-training phase as suggested in~\citet{ddpgfd}. We also use the Q-filter from~\citet{overcoming_exploration}. We train with a similar number of data points as in IL and \vsf for a reasonable comparison. We design a reward function for the smoothing task that, at each time step, provides reward equal to the change in coverage between two consecutive states. Inspired by~\citet{openai-dactyl}, we provide a $+5$ bonus for triggering a coverage success, and $-5$ penalty for pulling the fabric out of bounds.

\subsubsection{Smoothing and Folding Results with \vsfrss}

\begin{table}[t]
\caption{
Simulated smoothing experimental results for \vsfrss and the baselines in Section~\ref{ssec:policy}. We report final coverage and number of actions per episode, averaged over 200 simulated episodes per tier, and use the same random seeds for a fair comparison. \vsfrss performs well even for difficult starting states. It attains similar final coverage as the \il (IL) agent from~\cite{seita_ryan} and outperforms the other baselines. The \vsfrss and IL agents were trained on equal amounts of domain-randomized RGBD data, but the IL agent has a demonstrator for every training state, whereas \vsfrss is trained with data collected from a random policy.
}
\centering
\begin{tabular}{l l | r r}
\textbf{Tier} & \textbf{Method} & \textbf{Coverage} & \textbf{Actions} \\ \hline 
1 & Random  &  25.0 $\pm$ 14.6 & 2.4 $\pm$ 2.2  \\
1 & Highest &  66.2 $\pm$ 25.1 & 8.2 $\pm$ 3.2  \\
1 & Wrinkle &  91.3 $\pm$ \:\:7.1  & 5.4 $\pm$ 3.7  \\
1 & DDPG and Demos & 87.1 $\pm$ 10.7 & 8.7 $\pm$ 6.1 \\
1 & \il & 94.3 $\pm$ \:\:2.3  & 3.3 $\pm$ 3.1 \\
1 & \vsfrss & 92.5 $\pm$ \:\:2.5 & 8.3 $\pm$ 4.7 \\ \hline
2 & Random  &  22.3 $\pm$ 12.7  & 3.0 $\pm$ 2.5  \\ 
2 & Highest &  57.3 $\pm$ 13.0  & 10.0 $\pm$ 0.3  \\
2 & Wrinkle &  87.0 $\pm$ 10.8  & 7.6 $\pm$ 2.8  \\
2 & DDPG and Demos & 82.0 $\pm$ 14.7 & 9.5 $\pm$ 5.8 \\
2 & \il &  92.8 $\pm$ \:\:7.0  &  5.7 $\pm$ 4.0 \\
2 & \vsfrss & 90.3 $\pm$ \:\:3.8 & 12.1 $\pm$ 3.4 \\ \hline
3 & Random  &  20.6 $\pm$ 12.3  & 3.8 $\pm$ 2.8  \\ 
3 & Highest &  36.3 $\pm$ 16.3  & 7.9 $\pm$ 3.2  \\ 
3 & Wrinkle &  73.6 $\pm$ 19.0  & 8.9 $\pm$ 2.0  \\ 
3 & DDPG and Demos & 67.9 $\pm$ 15.6 & 12.9 $\pm$ 3.9 \\
3 & \il &  88.6 $\pm$ 11.5 & 10.1 $\pm$ 3.9  \\
3 & \vsfrss & 89.3 $\pm$ \:\:5.9 & 13.1 $\pm$ 2.9 \\
\end{tabular}
\vspace*{-5pt}
\label{tab:analytic}
\end{table}

Results in Table~\ref{tab:analytic} indicate that \vsfrss significantly outperforms the analytic and model-free reinforcement learning baselines for fabric smoothing in simulation. It has similar performance to the IL agent, a ``smoothing specialist'' that rivals the performance of the corner pulling demonstrator used in training (see Appendix~\ref{app:implementation}). See Figure~\ref{fig:rollout_smooth} for an example Tier 3 \vsfrss episode. Furthermore, we find that coverage values are statistically significant compared to all baselines other than IL, and that performance is not notably impacted by the use of domain randomization. Results from the Mann-Whitney U test~\cite{mannwhitney} and a domain randomization ablation study are reported in Appendix~\ref{app:results}. \vsfrss, however, requires more actions than DAgger, especially on Tier 1, with 8.3 actions per episode compared to 3.3 per episode. Attempting to mitigate this by increasing the variance of CEM results in poor performance, as actions are sampled outside the truncated action distribution used to generate data. 
Indeed, this is one of the motivations for the design of \dnew, as described in Section~\ref{sssec:new-data-gen}.

\begin{figure}[t]
\center
\includegraphics[width=0.50\textwidth]{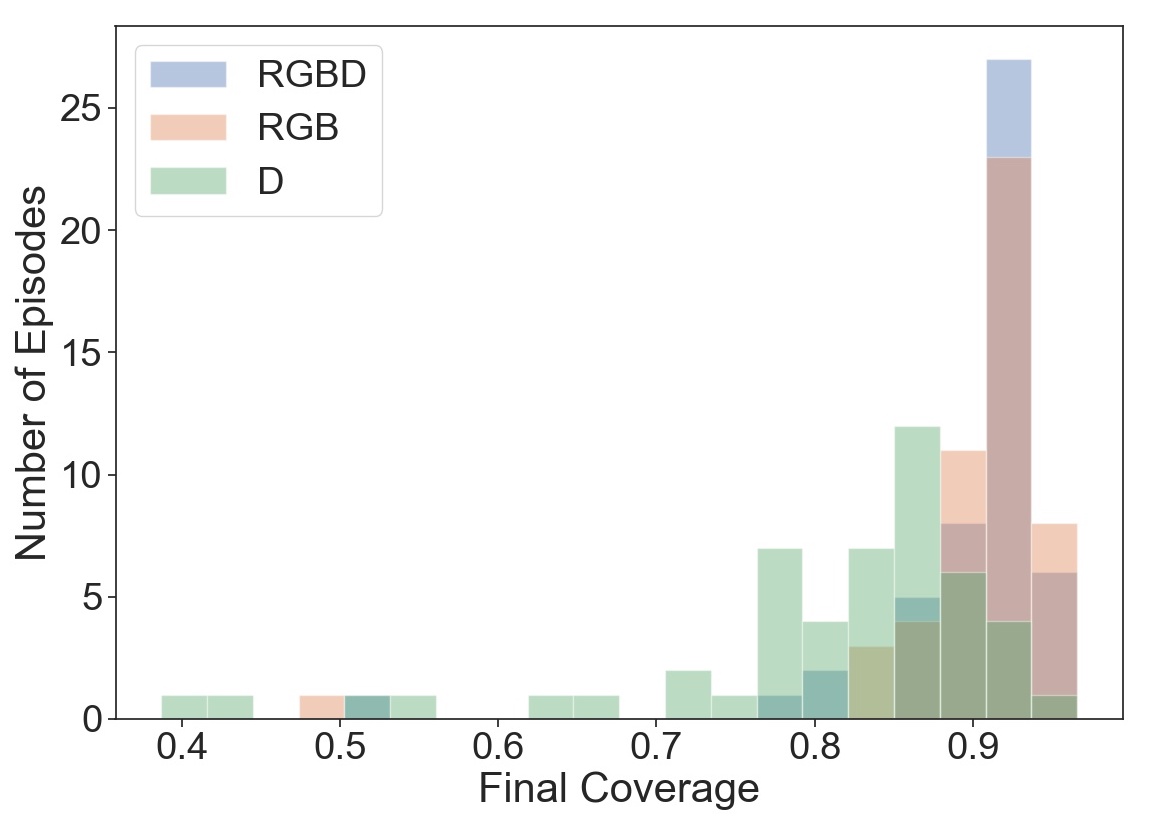}
\caption{
Final coverage values on 50 \vsfrss simulated smoothing episodes from Tier 3 starting states. We fix the random seed so that each input modality (RGB, D and RGBD) begins with the same starting states.
}
\vspace*{-5pt}
\label{fig:hist}
\end{figure}

\begin{table}[t]
\caption{
Simulated single folding (1-Fold) results. We run \vsfrss with the goal image in Figure~\ref{fig:rollout_folding} for 20 episodes when L2 is taken on the depth, RGB, and RGBD channels. The results suggest that adding depth allows us to significantly outperform RGB-only Visual Foresight on this task.
}
\centering
\begin{tabular}{l | r r r}
\textbf{Cost Function} & \textbf{Successes} & \textbf{Failures} & \textbf{\% Success} \\ \hline 
L2 Depth  & 0 & 20 & 0\%  \\
L2 RGB  & 10 & 10 & 50\%  \\
\textbf{L2 RGBD}  & \textbf{18} & \textbf{2} & \textbf{90\%}  \\
\end{tabular}
\vspace*{-5pt}
\label{tab:folding_trials}
\end{table}

\begin{figure}[t]
\center
\includegraphics[width=0.8\textwidth]{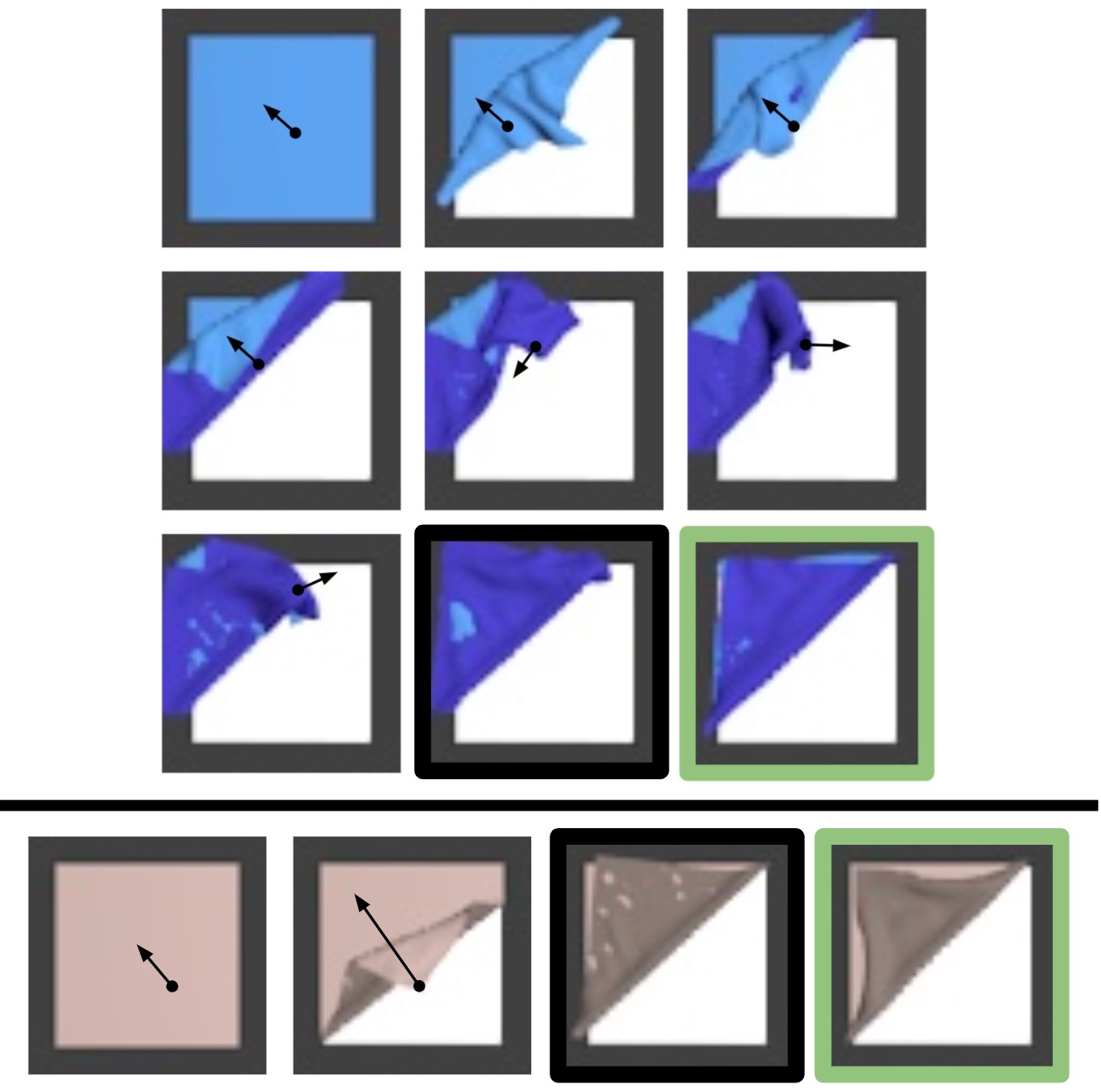}
\caption{
RGB observations of successful folding episodes in simulation. \textbf{Top:} A rollout using \vsfrss. The goal image is boxed in green, while the final frame in the episode is boxed in black. Here it takes 7 actions (left-to-right, top-to-bottom) from smooth to approximately folded. \textbf{Bottom:} A rollout using \vsf trained on the new dataset \dnew (Section~\ref{ssec:auro-sim}). Here it only takes 2 actions and results in a higher quality fold. There are several areas of the fabric simulator which have overlapping layers due to the difficulty of accurately modeling fabric-fabric collisions in simulation, which explain the light patches in the figure.
}
\vspace*{-10pt}
\label{fig:rollout_folding}
\end{figure}


We proceed to study the effect of the input modality (i.e. RGB, D, and RGBD) in \vsfrss. See Figure~\ref{fig:hist} for a histogram of coverage values obtained on the simulated smoothing task. Here RGBD performs the best but only slightly outperforms RGB, which is perhaps unsurprising due to the relatively low depth variation in the smoothing task. We also vary the input modality in a fabric \textit{folding} task. For folding, we use the same video prediction model, trained only with random interaction data, and keep planning parameters the same besides the initial CEM variance (see Appendix~\ref{app:visualmpc}). We change the goal image to the triangular, folded shape shown in Figure~\ref{fig:rollout_folding} and change the initial state to a smooth state (which can be interpreted as the result of smoothing). The two sides of the fabric are shaded differently, with the darker shade on the bottom layer. Due to the action space bounds (Section~\ref{ssec:data-gen}), getting to this goal state directly is not possible in less than two actions and requires a precise sequence of pick-and-pull actions. 

We visually inspect the final states in each episode, and classify them as successes or failures, as done in other work on fabric folding~\cite{lee2020learning}. For RGBD images, this decision boundary empirically corresponds to an L2 threshold of about 8000; see Figure~\ref{fig:rollout_folding} for a typical success case. In Table~\ref{tab:folding_trials} we compare performance of L2 cost taken over RGB, depth, and RGBD channels. RGBD significantly outperforms the other modes, which correspond to Visual Foresight and ``Spatial Foresight'' (depth only) respectively, suggesting the usefulness of augmenting Visual Foresight with depth maps.

\subsection{Simulation Results from Variations of VSF Settings}\label{ssec:auro-sim}

This section contains newer results not in prior work~\cite{vsf-fabric}. Here, we study the choice of dataset, visual dynamics model, optimization method, and planning cost function on performance in simulation. We evaluate 20 trials of smoothing, folding (``1-Fold''), and double folding (``2-Fold''), on each of 12 possible settings. See Figure~\ref{fig:teaser} for examples of these goals. Note that ``1-Fold" refers to the structure of the \textit{goal image}, not the minimum number of actions it requires to reach (which is 2 with the current action space). Recall that \vsfrss in prior work~\cite{vsf-fabric} and the previous section represents 1 of these 12 settings (namely, \dold, SV2P, CEM, and Pixel L2) for \vsf. Note that we consider 12 settings instead of 16 because we did not record the fabric mesh state when generating \dold, which makes combinations of \dold and the learned cost function from Equation~\ref{eq:cost2} impossible. See Table~\ref{tab:auro-sim} for quantitative results, which contains all $3 \times 12 = 36$ combinations of choices for VSF. We find that no single combination achieves the best performance on all three tasks, suggesting tradeoffs in the selection of each component of VSF.

\begin{table}[t]
\small
\caption{
Success rate (or mean coverage for smoothing) and number of actions (among successful episodes, or ``N/A'' if no successful episodes) of smoothing (``Smooth"), single folding (``1-Fold"), and double folding (``2-Fold") on all possible settings of dataset, visual dynamics model, optimization method, and planning cost (see Section~\ref{sec:approach}). We run 20 trials for each row. All smoothing results are from Tier 3 starting states.
}
\centering
\begin{tabular}{l l l l l l | r r}
& \textbf{Dataset} & \textbf{Model} & \textbf{Optimizer} & \textbf{Cost} & \textbf{Task} & \textbf{Success} & \textbf{\# Actions} \\ \hline 
1 & \dold & SV2P & CEM & Pixel L2 & Smooth & 86.5 & 10.7 $\pm$ 4.1 \\
2 & \dold & SV2P & CMA-ES & Pixel L2 & Smooth & 50.0 & 5.0 $\pm$ 4.0 \\
3 & \dold & SVG & CEM    & Pixel L2 & Smooth & 71.3 & 10.8 $\pm$ 3.9  \\
4 & \dold & SVG & CMA-ES & Pixel L2 & Smooth & 41.0 & 5.2 $\pm$ 4.1 \\
5 & \dnew & SV2P & CEM & Pixel L2 & Smooth & 84.4 & 12.4 $\pm$ 3.9 \\
6 & \dnew & SV2P & CEM & Vertex L2 & Smooth & \bf{88.0} & 10.9 $\pm$ 3.9 \\
7 & \dnew & SV2P & CMA-ES & Pixel L2 & Smooth & 44.2 & 6.4 $\pm$ 4.7 \\
8 & \dnew & SV2P & CMA-ES & Vertex L2 & Smooth & 48.3 & 6.8 $\pm$ 5.6 \\
9 & \dnew & SVG & CEM & Pixel L2 & Smooth & 67.0 & 12.3 $\pm$ 3.6 \\
10 & \dnew & SVG & CEM & Vertex L2    & Smooth & 72.8 & 10.1 $\pm$ 4.0 \\
11 & \dnew & SVG & CMA-ES & Pixel L2  & Smooth & 39.3 & 6.8 $\pm$ 4.7 \\
12 & \dnew & SVG & CMA-ES & Vertex L2 & Smooth & 44.3 & 6.7 $\pm$ 4.8 \\
\hline
13 & \dold & SV2P & CEM & Pixel L2 & 1-Fold & 90 & 8.3 $\pm$ 1.2 \\
14 & \dold & SV2P & CMA-ES & Pixel L2 & 1-Fold & 5 & 6.0 $\pm$ 0.0 \\
15 & \dold & SVG & CEM    & Pixel L2 & 1-Fold & 0 & N/A \\
16 & \dold & SVG & CMA-ES & Pixel L2 & 1-Fold & 0 & N/A \\
17 & \dnew & SV2P & CEM & Pixel L2 & 1-Fold & \bf{95} & 2.0 $\pm$ 0.0 \\
18 & \dnew & SV2P & CEM & Vertex L2 & 1-Fold & 90 & 2.1 $\pm$ 0.2 \\
19 & \dnew & SV2P & CMA-ES & Pixel L2 & 1-Fold & 15 & 1.3 $\pm$ 0.5 \\
20 & \dnew & SV2P & CMA-ES & Vertex L2 & 1-Fold & 10 & 3.0 $\pm$ 2.0 \\
21 & \dnew & SVG & CEM    & Pixel L2  & 1-Fold & 10 & 8.5 $\pm$ 2.1 \\
22 & \dnew & SVG & CEM    & Vertex L2 & 1-Fold & 10 & 2.5 $\pm$ 0.7 \\
23 & \dnew & SVG & CMA-ES & Pixel L2  & 1-Fold &  0 &  N/A \\
24 & \dnew & SVG & CMA-ES & Vertex L2 & 1-Fold &  0 &  N/A \\
\hline
25 & \dold & SV2P & CEM & Pixel L2 & 2-Fold & 30 & 5.2 $\pm$ 1.7 \\
26 & \dold & SV2P & CMA-ES & Pixel L2 & 2-Fold & 30 & 3.3 $\pm$ 0.9 \\
27 & \dold & SVG & CEM & Pixel L2    & 2-Fold & 0 & N/A \\
28 & \dold & SVG & CMA-ES & Pixel L2 & 2-Fold & 0 & N/A \\
29 & \dnew & SV2P & CEM & Pixel L2 & 2-Fold & 10 & 7.5 $\pm$ 2.5 \\
30 & \dnew & SV2P & CEM & Vertex L2 & 2-Fold & 10 & 5.5 $\pm$ 0.5 \\
31 & \dnew & SV2P & CMA-ES & Pixel L2 & 2-Fold & 15 & 3.3 $\pm$ 1.3 \\
32 & \dnew & SV2P & CMA-ES & Vertex L2 & 2-Fold & \bf{40} & 2.4 $\pm$ 0.5 \\
33 & \dnew & SVG & CEM    & Pixel L2  & 2-Fold & 0 & N/A \\
34 & \dnew & SVG & CEM    & Vertex L2 & 2-Fold & 5 & 8.0 $\pm$ 0.0 \\
35 & \dnew & SVG & CMA-ES & Pixel L2  & 2-Fold & 0 & N/A \\
36 & \dnew & SVG & CMA-ES & Vertex L2 & 2-Fold & 0 & N/A 
\end{tabular}
\vspace*{-10pt}
\label{tab:auro-sim}
\end{table}

\subsubsection{Dataset Comparison}
Keeping all \vsfrss settings constant besides dataset choice indicates that \dnew has no noticeable impact on smoothing performance (Row 5), improves folding performance (Row 17), and hurts double folding performance (Row 29). However, the best setting for double folding performance includes \dnew (Row 32). The most dramatic improvement from \dnew is in the efficiency of the folding rollouts. In particular, with SV2P, CEM, and Pixel L2, switching the dataset alone decreases the mean number of actions from 8.3 to 2.0 and yields higher quality rollouts (see Figure~\ref{fig:rollout_folding}). This increase in efficiency suggests that physical fabric folding will be more feasible, as the sim-to-real dynamics mismatch is not able to compound over time (Section~\ref{ssec:physical-folding}). 

\subsubsection{Visual Dynamics Model Comparison}
In all smoothing and folding experiments, results suggest that planning using the action-conditioned SVG video prediction model from Figure~\ref{fig:svg-action} leads to lower quality results compared to SV2P. The best smoothing result with SVG (row 10, 72.8\% coverage, with CEM and Vertex L2 on \dnew) lags behind the best SV2P smoothing result (row 6, 88.0\% coverage, also with CEM and Vertex L2 on \dnew). We hypothesize that the lower performance metrics with SVG as compared to SV2P may be partially explained from the model architectures plus the amount of data available. The architecture of SV2P~\cite{sv2p}, based on~\cite{finn_cdna}, involves predicting transformations of pixels which are constrained to avoid moving too much in predicted future images, whereas SVG~\cite{svg} does not apply a similar constraint when predicting images. This may cause it to predict images of highly disjoint fabric, which we qualitatively observe in the image predictions during MPC. While the constraints imposed on SV2P may cause it to be less expressive than SVG given sufficient data, the data size of \dnew, containing 99,320 data transitions (see Section~\ref{ssec:data-gen}) may not be large enough to show benefits for SVG.

\subsubsection{Optimization Method Comparison}
\begin{figure}[t]
\center
\includegraphics[width=0.95\textwidth]{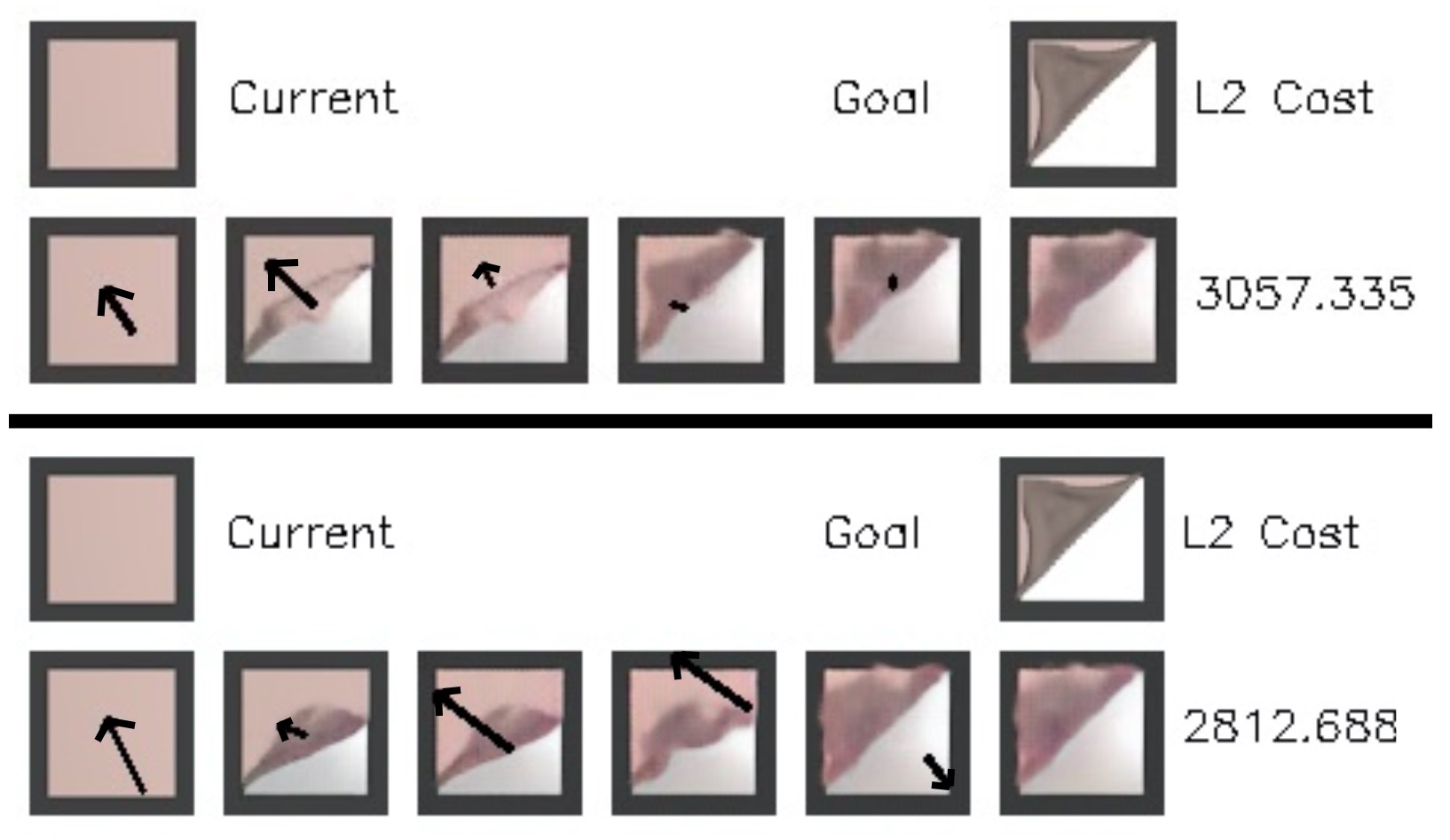}
\caption{
\textbf{Top:} A VSF folding plan with \dnew, SV2P, Pixel L2 cost and the CEM optimizer. \textbf{Bottom:} A VSF folding plan with \dnew, SV2P, Pixel L2 cost and the CMA-ES optimizer. As in Figure~\ref{fig:planning}, the five images after the current image are generated by the visual dynamics model. CMA-ES arrives at a lower cost solution but converges on more drastic actions that are more liable to result in states that deviate from predictions, and may cause the fabric to go out of bounds.
}
\label{fig:cem-cma-1}
\end{figure}

In all smoothing and folding experiments involving CMA-ES, performance is far below that of CEM. However, CMA-ES improves performance and efficiency in double folding, especially among the experiments involving \dnew. To better understand why this is the case, we inspect VSF plans for single folding in Figure~\ref{fig:cem-cma-1} and double folding in Figure~\ref{fig:cem-cma-2}. Despite the poor performance on folding, we find that CMA-ES actually arrives at a lower cost solution, indicating that CMA-ES may be exploiting inaccuracies in the visual dynamics model. The CMA-ES solution generally involves larger action deltas that can cause resulting states to deviate from the predictions, which may be due to the smaller population size during optimization. However, for the double-folding experiments (Figure~\ref{fig:cem-cma-2}), CEM is unable to find a high quality solution, while CMA-ES is able to find one. We hypothesize that this behavior is due to averaging over a multimodal optimization landscape with a large population size. Due to the structure of the double folding goal image, the order in which the top right corner and bottom left corner are folded toward the center does not impact the quality of the solution. Since we run CMA-ES with many fewer samples per iteration, it is less likely to reach both optima at the same time.

\subsubsection{Cost Function Comparison}
\begin{figure}[t]
\center
\includegraphics[width=0.95\textwidth]{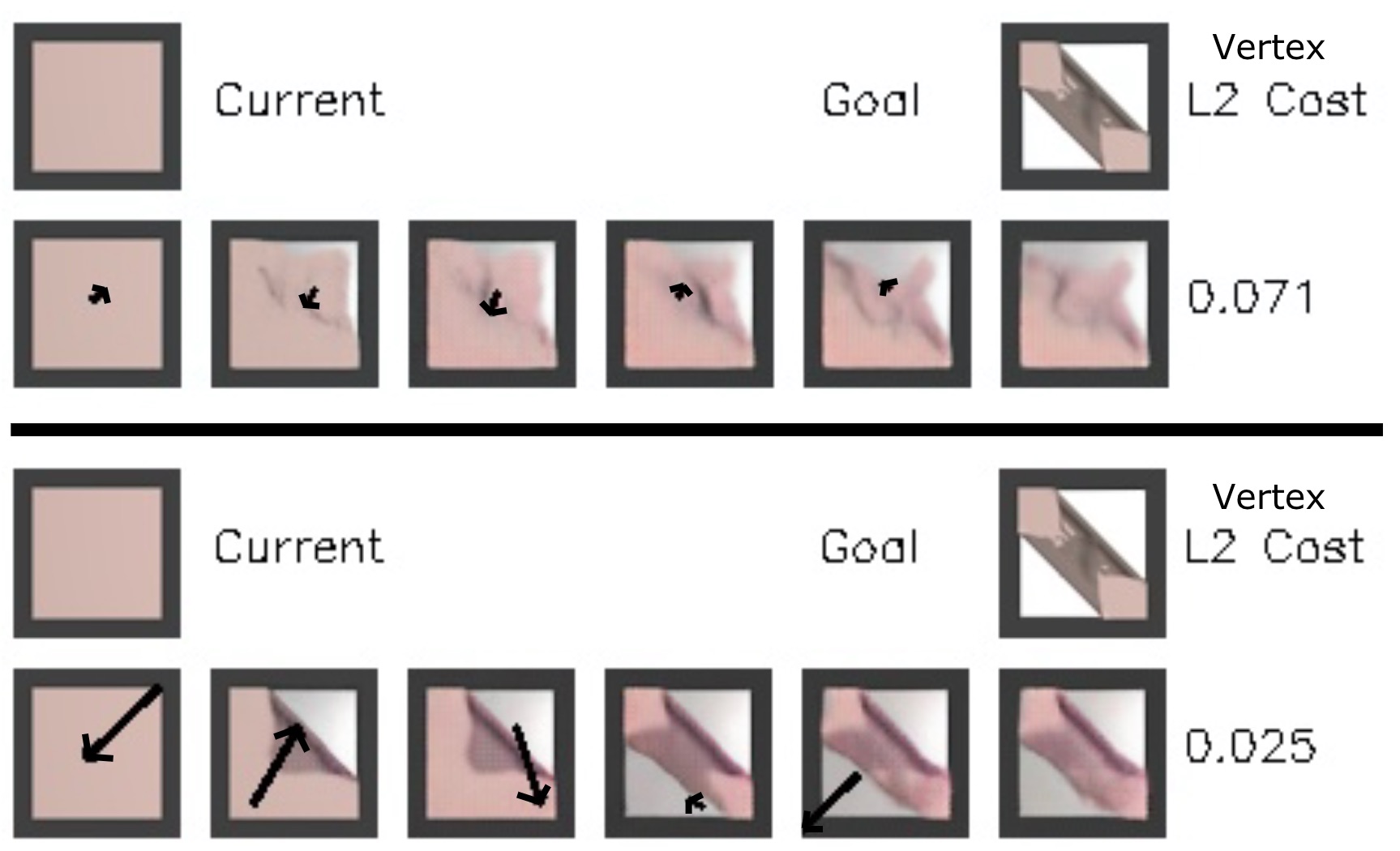}
\caption{
\textbf{Top:} A VSF double folding plan with \dnew, SV2P, Vertex L2 cost and the CEM optimizer. \textbf{Bottom:} A VSF double folding plan with \dnew, SV2P, Vertex L2 cost and the CMA-ES optimizer. CMA-ES is able to find a much better plan for achieving the goal state.
}
\label{fig:cem-cma-2}
\end{figure}

In smoothing and folding experiments, changing the cost function to the learned vertex distance estimator does not have significant impact on performance in either direction, though Vertex L2 does slightly boost coverage for smoothing. This is perhaps unsurprising, as Pixel L2 is likely sufficient for goal images with simple visual structure (i.e., a square or triangle with a single color). However, with the more complex double folding goal image, comparison of Rows 31 and 32 (where dataset is \dnew, visual dynamics model is SV2P, and optimizer is CMA-ES) indicates that the learned Vertex L2 significantly outperforms Pixel L2. In Figure~\ref{fig:cem-cma-2} we see that minimizing the Vertex L2 cost appropriately guides CMA-ES to a trajectory with a final predicted image similar to the double folding goal image.

%% file: 6-physical-exps.tex
\section{Physical Experiments}\label{sec:physical-results}

We evaluate \vsf on a physical da Vinci surgical robot~\cite{dvrk2014}. We use the same experimental setup as in~\citet{seita_ryan}, including the calibration procedure to map pick points $(x,y)$ into positions and orientations with respect to the robot's base frame. The sequential tasks we consider are challenging due to the robot's imprecision~\cite{seita_icra_2018}. We use a Zivid One Plus camera mounted 0.9 meters above the workspace to capture RGBD images. We manipulate a 5" by 5" piece of fabric (blue for smoothing, brown for folding) and apply some damping to mitigate stiffness due to its small size. In Section~\ref{ssec:physical-smoothing}, we report results from our prior work~\cite{vsf-fabric}. In Section~\ref{ssec:physical-folding}, we show new results with physical fabric folding using a set of parameters which we refer to as ``\vsfnew.''

\begin{table}[t]
\caption{
Physical smoothing robot experiment results for \il (IL), i.e. DAgger, and \vsfrss. For both methods, we choose the policy snapshot with highest performance in simulation, and each are applied on all tiers (T1, T2, T3). We show results across 10 episodes of IL per tier and 5 episodes of \vsfrss per tier, and show average starting and final coverage, maximum coverage at any point per episode, and the number of actions. Results suggest that \vsfrss attains final coverage comparable to or exceeding that of IL despite not being trained on demonstration data, though \vsfrss requires more actions per episode.
}
\centering
\begin{tabular}{l | l l l r }
(Tier) Method & (1) Start & (2) Final & (3) Max & (4) Actions \\ \hline 
(1) IL & 74.2 $\pm$ 5 & 92.1 $\pm$ \:\:6 & 92.9 $\pm$ \:\:3 & 4.0 $\pm$ 3 \\
(1) \vsfrss & 78.3 $\pm$ 6 & \textbf{93.4 $\pm$ \:\:2} & 93.4 $\pm$ \:\:2 & 8.2 $\pm$ 4 \\ \hline
(2) IL & 58.2 $\pm$ 3 & 84.2 $\pm$ 18 & 86.8 $\pm$ 15 & 9.8 $\pm$ 5 \\
(2) \vsfrss & 59.5 $\pm$ 3 & \textbf{87.1 $\pm$ \:\:9} & 90.0 $\pm$ \:\:5 & 12.8 $\pm$ 3 \\ \hline
(3) IL & 43.3 $\pm$ 4 & 75.2 $\pm$ 18 & 79.1 $\pm$ 14 & 12.5 $\pm$ 4 \\
(3) \vsfrss & 41.4 $\pm$ 3 & \textbf{75.6 $\pm$ 15} & 76.9 $\pm$ 15 & 15.0 $\pm$ 0 \\
\end{tabular}
\vspace*{-5pt}
\label{tab:surgical}
\end{table}

\begin{figure*}[t]
\center
\includegraphics[width=1.00\textwidth]{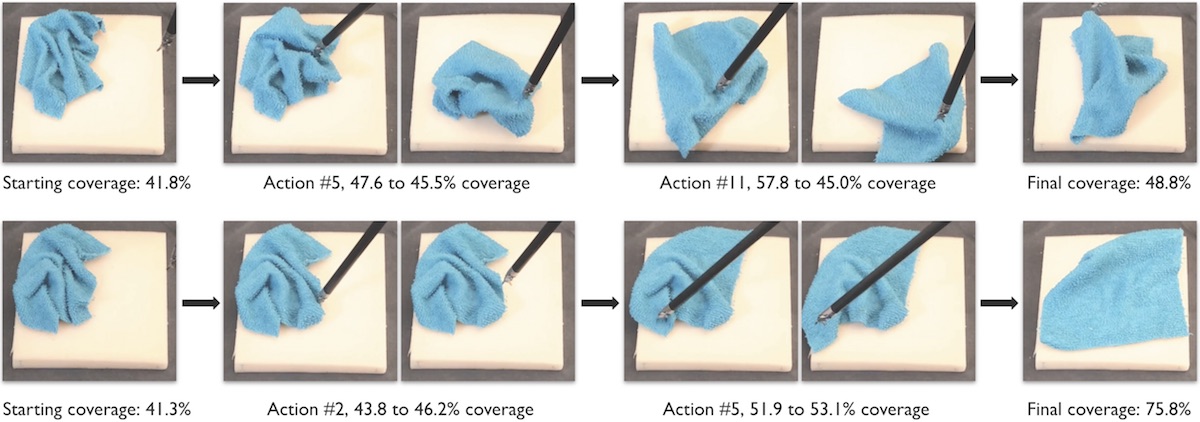}
\caption{
A qualitative comparison of physical da Vinci episodes with an \il policy (top row) and a \vsf policy (bottom row) from our prior work~\cite{vsf-fabric} using \vsfrss settings. The rows show screen captures taken from the third-person video view for recording episodes; these are not the input to \vsf. To facilitate comparisons among IL and \vsfrss, we manually make the starting fabric state as similar as possible. Over the course of several actions, the IL policy sometimes takes actions that are highly counter-productive, such as the 5th and 11th actions above. Both pick points are reasonably chosen, but the large deltas cause the lower right fabric corner to get hidden. In contrast, \vsfrss takes shorter pulls on average, with representative examples shown above for the 2nd and 5th actions. At the end, the IL policy gets just 48.8\% coverage (far below its usual performance), whereas \vsfrss gets 75.8\%. See Table~\ref{tab:surgical} for more results.
}
\vspace*{-5pt}
\label{fig:il_vs_vf_example}
\end{figure*}

\subsection{Physical Fabric Smoothing}\label{ssec:physical-smoothing}

Results in this section are from our prior work~\cite{vsf-fabric}. We evaluate the \il and \vsfrss policies from Section~\ref{ssec:policy}. We do not test with the model-free DDPG policy baseline, as it performed significantly worse than the other two methods. For IL, this is the final model trained with 110,000 actions based on a corner-pulling demonstrator with access to state information. This uses slightly more than the 105,045 actions used for training \vsfrss. To match the simulation setup, we limit each episode to a maximum of 15 actions. For both methods, we initialize the fabric in highly rumpled states which mirror those from the simulated tiers. We run ten episodes per tier for IL and five episodes per tier for \vsfrss, for 45 episodes in all. Within each tier, we attempt to make starting fabric states reasonably comparable among IL and \vsfrss episodes (see Figure~\ref{fig:il_vs_vf_example}). We present quantitative results in Table~\ref{tab:surgical} that suggest that \vsfrss gets final coverage results comparable to that of IL, despite not being trained on any corner-pulling demonstration data. However, it sometimes requires more actions to complete an episode and takes significantly more time to plan an action (on the order of 20 more seconds per action), since the Cross Entropy Method requires thousands of forward passes through a deep neural network while IL requires only a single pass.


As an example, Figure~\ref{fig:il_vs_vf_example} shows a time lapse of a subset of actions for one episode from IL and \vsfrss. Both begin with a fabric of roughly the same shape to facilitate comparisons. On the fifth action, the IL policy has a pick point that is slightly north of the ideal spot. The pull direction to the lower right fabric plane corner is reasonable, but due to the pull length, combined with a slightly suboptimal pick point, the lower right fabric corner gets covered. This makes it harder for a policy trained from a corner-pulling demonstrator to get high coverage. In contrast, the \vsfrss policy takes actions of shorter magnitudes and does not fall into this trap.

\subsection{Physical Fabric Folding}\label{ssec:physical-folding}

\begin{figure}[t]
\center
\includegraphics[width=0.70\textwidth]{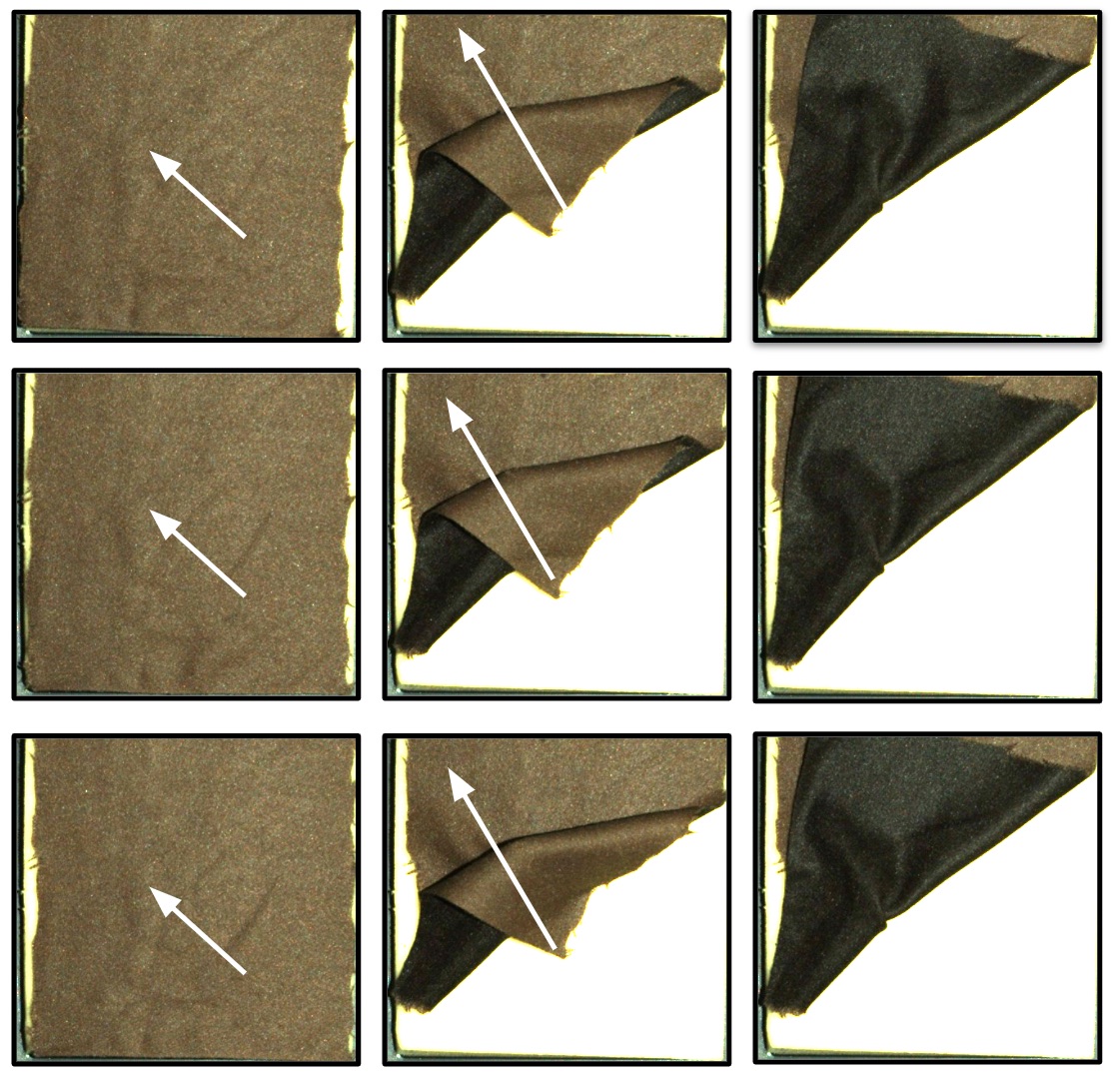}
\caption{
We evaluate the \vsfnew policy on a fabric folding task on the physical surgical robotic system with \dnew, SV2P, CEM, and Pixel L2 cost. Each of the three rows depicts top-down RGB images of one of the 9 out of 10 successful 2-step rollouts with a square piece of brown fabric. 
}
\vspace*{-10pt}
\label{fig:real-folding}
\end{figure}

We next evaluate \vsf on a fabric folding task, starting from a smooth state. In our prior work~\cite{vsf-fabric}, we were unable to successfully perform folding on the physical system with \vsfrss. When comparing the real fabric with simulated fabric, we found a gap between the physics of our simulator and that of the real fabric, as fabric dynamics are notoriously difficult to model~\cite{arcsim2012}. Unlike smoothing, folding can require more nuanced actions such as reversing the surface normals of the fabric. Since \vsfrss took 8.3 actions on average to fold the fabric in simulation, in real episodes this dynamics gap compounded over time and was exacerbated by the imprecision of cable-driven robots like the dVRK~\cite{seita_icra_2018}. In this work, we find that training visual dynamics on \dnew makes it possible for \vsf to successfully fold fabric in simulation with just 2 actions in 19 out of the 20 successful trials (Row 17 in Table~\ref{tab:auro-sim}), which may be short enough to prevent the dynamics gap from building to irrecoverable levels. We refer to this new set of VSF parameters using \dnew, SV2P, CEM, and Pixel L2 as ``\vsfnew.''

To test this hypothesis, we perform a VSF-2.0 plan on the physical system in an open-loop fashion. Such an approach is viable only if it is possible to register the initial fabric state into simulation; in the fabric folding case, this is trivial, as the fabric starts fully smooth. To correct for near misses, the system moves the pick point to the nearest point on the fabric, which it computes by color masking the real RGB observations. In 9 of 10 trials, the robot successfully folds using two actions, with the only failure case due to picking multiple layers of the fabric when intending to pick the top layer. See Figure~\ref{fig:real-folding} for 3 of these trials and the project website for videos.

%% file: 7-conclusion.tex
\section{Conclusion and Future Work}\label{sec:conclusions}

Our prior work~\cite{vsf-fabric} presented \vsf, which leverages a combination of RGB and depth information to learn goal conditioned fabric manipulation policies for a variety of sequential tasks. In~\cite{vsf-fabric}, we train a video prediction model on purely random interaction data with fabric in simulation, and demonstrate that planning over this model with MPC results in a policy that achieves $90\%$ success rate for fabric smoothing and folding tasks.

In this work, we investigate new alternatives to the four core aspects of \vsf: data generation, visuospatial dynamics model, cost function, and optimization procedure. To improve data, we introduce \dnew as a new dataset with longer action pull vector magnitudes and a bias towards picking at corners. We propose and test an action-conditioned version of SVG for modeling visuospatial dynamics. To optimize the MPC objective during planning, we test Covariance Matrix Adaptation Evolution Strategy (CMA-ES). Finally, we train a learned cost function as an alternative to L2 pixel differences in images. Smoothing and folding results in simulation suggest that the new data, \dnew, is the most promising route to improving results. Using this new data for \vsf allows us to learn more accurate visual dynamics models because the dataset contains actions that are more broadly relevant for fabric manipulation. These actions are biased towards picking fabric corners and have magnitudes more reflective of the actions required to do fabric manipulation tasks such as smoothing and folding. The resulting improvement in efficiency led to successful fabric folding on the physical robotic system in 9 out of 10 trials, while in our prior work~\cite{vsf-fabric} we were unable to successfully fold fabric in physical trials.

In light of these results, future work will attempt to understand the effect of the distribution and magnitude of the dataset used to train visual dynamics models on \vsf. We plan to generate orders of magnitude more data, and will benchmark performance as a function of data size and other properties. In addition, we will test \vsf on different fabric shapes, and investigate ways to incorporate bilateral manipulation or human-in-the-loop policies.

%% file: 8-appendix.tex
\clearpage
\normalsize

We structure this Appendix as follows:

\begin{itemize}
    \item Appendix~\ref{app:simulators} compares and contrasts various fabric simulators.
    \item Appendix~\ref{app:implementation} lists hyperparameters and provides details for training policies.
    \item Appendix~\ref{app:results} provides more details on the smoothing experiments.
\end{itemize}

\section{Fabric Simulators}\label{app:simulators}

As in the prior paper~\cite{vsf-fabric}, we use the fabric simulator originally developed in~\citet{seita_ryan}. This simulator possesses an ideal balance between ease of code implementation, speed, and accuracy, and was able to lead to reasonable smoothing policies in prior work. We considered using simulators from ARCSim~\cite{arcsim2012}, MuJoCo~\cite{mujoco}, PyBullet~\cite{coumans2019}, Blender~\cite{blender}, or NVIDIA FLeX~\cite{lin2020softgym},  but did not use them for several reasons outlined below. 

High-fidelity simulators, such as ARCSim, take too long to simulate to get sufficient data for training visual dynamics models. Furthermore, it is difficult to simulate rudimentary grasping behavior in ARCSim because it does not represent fabric as a fixed grid of vertices, which means grasping cannot be simulated by pinning vertices.

Blender includes a new fabric simulator, with substantial improvements after 2017 for more realistic shearing and tensioning. These changes, however, are only supported in Blender 2.8, not Blender 2.79, and we used 2.79 because Blender 2.8 does not allow background processes to run on headless servers, which prevented us from running mass data collection. Additionally, Blender does not allow the dynamic re-grasping of mesh vertices during simulation which makes long horizon cloth manipulation and data collection difficult. 

MuJoCo is a widely utilized physics simulator for deep reinforcement learning benchmarks~\cite{mujoco}. The first MuJoCo version providing full support for fabric manipulation was released in October 2018. Currently, the only work that integrates the fabric simulator with simulated robot grasps is from~\citet{lerrel}, which was developed concurrently with the prior work~\cite{vsf-fabric}. Upon investigating the open-source code, we found that MuJoCo's fabric simulator did not handle fabric self-collisions better than the simulator from~\cite{seita_ryan}, and hence did not pursue it further.

The PyBullet simulator code from~\citet{sim2real_deform_2018} showed relatively successful fabric simulation, but it was difficult for us to adapt the author's code to the proposed work, which made significant changes to the off-the-shelf PyBullet code. PyBullet's fabric simulator was upgraded and tested for more fabric-related tasks in~\citet{seita2020learning}, but still suffers from self-collisions and fabric which tends to get crumpled.

In concurrent work, SoftGym~\cite{lin2020softgym} benchmarks deep reinforcement learning algorithms on deformable object manipulation tasks, including those with fabrics. SoftGym provides fabric simulation environments utilizing NVIDIA FLeX, which models deformable objects in a particle and position based dynamical system similar to the mass-spring system used in the fabric simulator from~\cite{vsf-fabric,seita_ryan} and also incorporates self-collision handling. SoftGym is concurrent work, and future work will investigate the feasibility of utilizing Flex. Additionally, we will compare the performance of the model-based policies presented in this work to the model-free policies evaluated in~\cite{lin2020softgym} on similar smoothing and folding tasks.

%

\section{Details of Learning-Based Methods}\label{app:implementation}

\begin{figure}[t]
\center
\includegraphics[width=0.70\textwidth]{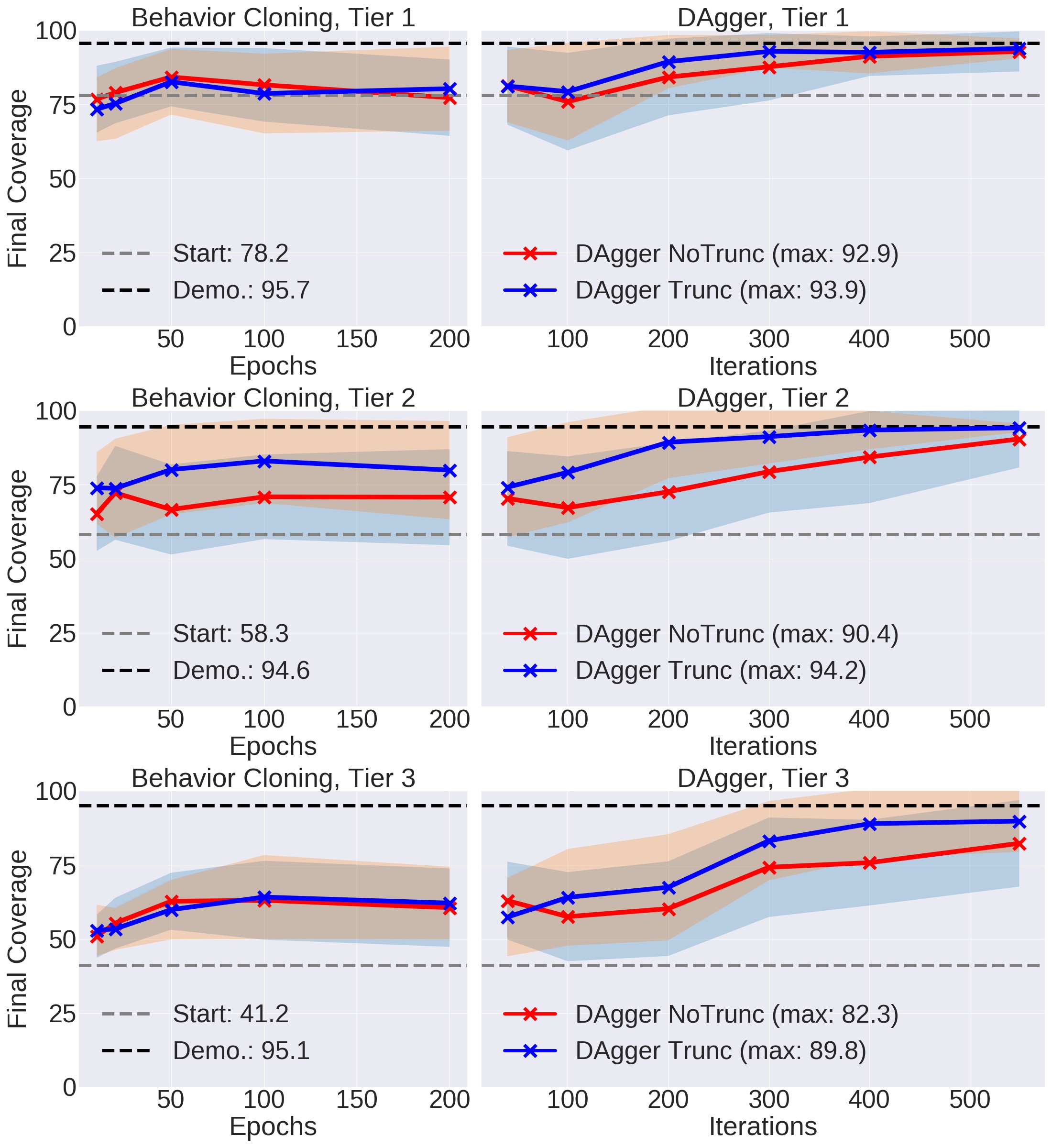}
\caption{
Average coverage over 50 simulated test-time episodes at checkpoints (marked ``X'') during the behavior cloning and DAgger phases. For each setting of no action truncation and action truncation, we deploy a single DAgger policy deployed on all tiers. Using dashed lines, we annotate the average starting coverage and the corner pulling demonstrator's average final coverage.
}
\label{fig:10-0-1}
\end{figure}

We describe implementation and training details of the three learning-based methods tested: imitation learning, model-free reinforcement learning, and model-based \vsf. The other baselines tested --- random, highest point, and wrinkles --- are borrowed unmodified from prior open-source code~\cite{seita_ryan}.

\subsection{Imitation Learning Baseline: DAgger}\label{app:dagger}

This section contains details and results from our prior work~\cite{vsf-fabric}. DAgger~\cite{ross2011reduction} is implemented directly from the open source DAgger code in~\citet{seita_ryan}. This was originally based on the open-source OpenAI baselines~\cite{baselines} library for parallel environment support to overcome the time bottleneck of fabric simulation.

We ran the corner pulling demonstrator for 2,000 trajectories, resulting in 6,697 image-action pairs $(\bo_t, \ba_t')$, where the notation $\ba_t'$ indicates the action is labeled and comes from the demonstrator. Each trajectory was randomly drawn from one of the three tiers in the simulator with equal probability. We then perform a behavior cloning~\cite{Pomerleau_behavior_cloning} ``pre-training'' period for 200 epochs over this offline data, which does not require environment interaction.

After behavior cloning, each DAgger iteration rolls out 20 parallel environments for 10 steps each (hence, 200 total new samples) which are labeled by the corner pulling policy, the same policy that created the offline data and uses underlying state information. These are inserted into a replay buffer of image-action samples, where all samples have actions labeled by the demonstrator. The replay buffer size is 50,000, but the original demonstrator data of size 6,697 is never removed from it. After environment stepping, we draw 240 minibatches of size 128 each for training and use Adam~\cite{adam2015} for optimization. The process repeats with the agent rolling out its updated policy. We run DAgger for 110,000 steps across all environments (hence, 5,500 steps per parallel environment) to make the number of actions consumed to be roughly the same as the number of actions used to train the video prediction model. This is significantly more than the 50,000 DAgger training steps in prior work~\cite{seita_ryan}. Table~\ref{tab:dagger-hyperparams} contains additional hyperparameters.

The actor (i.e., policy) neural network for DAgger uses a design based on~\citet{seita_ryan} and~\citet{sim2real_deform_2018}. The input to the policy are RGBD images of size $(56 \times 56 \times 4)$, where the four channels are formed from stacking an RGB and a single-channel depth image. The policy processes the input through four convolutional layers that have 32 filters with size $3\times 3$, and then uses four fully connected layers with 256 nodes each. The actor network has 0.8 million parameters.


The result from the actor policy is a 4D vector representing the action choice $\ba_t \in \mathbb{R}^4$ at each time step $t$. The last layer is a hyperbolic tangent which makes each of the four components of $\ba_t$ within $[-1,1]$. During action truncation, we further limit the two components of $\ba_t$ corresponding to the deltas into $[-0.4, 0.4]$.

A set of graphs representing learning progress for DAgger is shown in Figure~\ref{fig:10-0-1}, where for each marked snapshot, we roll it out in the environment for 50 episodes and measure final coverage. Results suggest the single DAgger policy, when trained with 110,000 total steps on RGBD images, performs well on all three tiers with performance nearly matching the 95-96\% coverage of the demonstrator.

We trained two variants of DAgger, one with and one without the action truncation to $[-0.4, 0.4]$ for the two deltas $\Delta x$ and $\Delta y$. The model trained on truncated actions outperforms the alternative setting, and it is also the setting used in \vsfrss, hence we use it for physical robot experiments. We choose the final snapshot as it has the highest test-time performance, and we use it as the policy for simulated and real benchmarks in the main part of the paper.

\begin{table}[t]
\caption{
DAgger hyperparameters. 
}
\centering
\begin{tabular}{l r}
\textbf{Hyperparameter} & \textbf{Value} \\  \hline
Parallel environments & 20 \\
Steps per env, between gradient updates & 10 \\
Gradient updates after parallel steps & 240 \\
Minibatch size            & 128 \\
Discount factor $\gamma$ & 0.99 \\
Demonstrator (offline) samples & 6697 \\
Policy learning rate   & 1e-4 \\
Policy $L_2$ regularization parameter  & 1e-5 \\
Behavior Cloning epochs & 200 \\
DAgger steps after Behavior Cloning    & 110000 \\
\end{tabular}
\label{tab:dagger-hyperparams}
\end{table}

\subsection{Model-Free Reinforcement Learning Baseline: DDPG}

\begin{figure}[t]
\center
\includegraphics[width=0.70\textwidth]{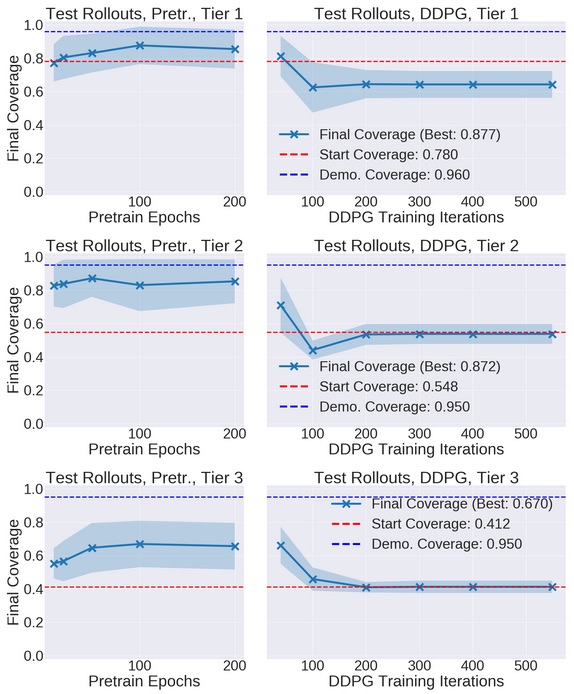}
\caption{
Average coverage over 50 simulated test-time episodes at checkpoints (marked ``X'') during the pre-training DDPG phase, and the DDPG phase with agent exploration. This is presented in a similar manner as in Figure~\ref{fig:10-0-1} for DAgger. Results suggest that DDPG has difficulty in training a policy that can achieve high coverage.
}
\label{fig:10-1-0}
\end{figure}

\begin{table}[t]
\caption{
DDPG hyperparameters. 
}
\centering
\begin{tabular}{l r}
\textbf{Hyperparameter} & \textbf{Value} \\  \hline
Parallel environments & 20 \\
Steps per env, between gradient updates & 10 \\
Gradient updates after parallel steps & 240 \\
Minibatch size            & 128 \\
Discount factor $\gamma$ & 0.99 \\
Demonstrator (offline) samples & 6697 \\
Actor learning rate   & 1e-4 \\
Actor $L_2$ regularization parameter  & 1e-5 \\
Critic learning rate   & 1e-3 \\
Critic $L_2$ regularization parameter  & 1e-5 \\
Pre-training epochs  & 200 \\
DDPG steps after pre-training & 110000 \\
\end{tabular}
\label{tab:ddpg-hyperparams}
\end{table}

This section contains details and results from our prior work~\cite{vsf-fabric}. To provide a second competitive baseline, we apply  model-free reinforcement learning. Specifically, we use a variant of Deep Deterministic Policy Gradients (DDPG)~\cite{ddpg2016} with several improvements as proposed in the research literature.  Briefly, DDPG is a deep reinforcement learning algorithm which trains parameterized actor and critic models, each of which are normally neural networks. The actor is the policy, and the critic is a value function.

First, as with DAgger, we use demonstrations~\cite{ddpgfd} to improve the performance of the learned policy. We use the same demonstrator data of 6,697 samples from DAgger, except this time each sample is a tuple of $(\bo_t, \ba_t', r_t, \bo_{t+1})$, including a scalar reward $r_t$ (to be described) and a successor state $\bo_{t+1}$. This data is added to the replay buffer and never removed. We use a pre-training phase (of 200 epochs) to initialize the actor and critic. We also apply $L_2$ regularization for both the actor and critic networks. In addition, we use the Q-filter from~\citet{overcoming_exploration} which may help the actor learn better actions than the demonstrator provides, perhaps for cases when naive corner pulling might not be ideal. For a fairer comparison, the actor network for DDPG uses the same architecture as the actor for DAgger. The critic has a similar architecture as the actor, with the only change that the action input $\ba_t$ is inserted and concatenated with the features of the image $\bo_t$ after the four convolutional layers, and before the fully connected portion. As with the imitation learning baseline, the inputs are RGBD images of size $(56\times 56 \times 4)$.

We design a dense reward to encourage the agent to achieve high coverage. At each time, the agent gets reward based on:

\begin{itemize}
    \item A small negative living reward of -0.05
    \item A small negative reward of -0.05 for failing to grasp any point on the fabric (i.e., a wasted grasp attempt).
    \item A delta in coverage based on the change in coverage from the current state and the prior state.
    \item A +5 bonus for triggering 92\% coverage.
    \item A -5 penalty for triggering an out-of-bounds condition, where the fabric significantly exceeds the boundaries of the underlying fabric plane.
\end{itemize}

We designed the reward function by informal tuning and borrowing ideas from the reward in~\citet{openai-dactyl}, which used a delta in joint angles and a similar bonus for moving a block towards a target, or a penalty for dropping it. Intuitively, an agent may learn to take a slightly counter-productive action which would decrease coverage (and thus the delta reward component is negative), but which may enable an easier subsequent action to trigger a high bonus. This reward design is only suited for smoothing. As with the imitation learning baseline, the model-free DDPG baseline is not designed for non-smoothing tasks.

Figure~\ref{fig:10-1-0} suggests that the pre-training phase, where the actor and critic are trained on the demonstrator data, helps increase coverage. The DDPG portion of training, however, results in performance collapse to achieving no net coverage. Upon further inspection, this is because the actions collapsed to having no ``deltas,'' so the robot reduces to picking up but then immediately releasing the fabric. Due to the weak performance of DDPG, we do not benchmark the policy on the physical robot.

\subsection{VisuoSpatial Foresight}\label{app:visualmpc}

\begin{table}[t]
\caption{
Visual MPC hyperparameters for CEM and CMA-ES. The notation $(\cdot) \times H$ indicates tiling the preceding array $H = 5$ times to fill the planning horizon.
}
\centering
\begin{tabular}{l r}
\textbf{Hyperparameter} & \textbf{Value} \\  \hline
Number of CEM iterations & 10 \\
CEM population size & 2000 \\
CEM $\alpha$ & 0.1 \\
CEM planning horizon & 5 \\
CEM initial mean $\mu$ & (0, 0, 0, 0) $ \times H$ \\
CEM initial variance $\Sigma$ & (0.25, 0.25, 0.08, 0.08) $ \times H$  \\
Number of CMA-ES iterations & 250 \\
CMA-ES population size & 12 \\
CMA-ES initial mean $\mu$ & (0, 0, 0, 0) $ \times H$  \\
CMA-ES initial variance $\Sigma$ & (0.25, 0.25, 0.25, 0.25) $ \times H$  \\
\end{tabular}
\label{tab:vismpc-hyperparams}
\end{table}

The main technique considered in this paper and our prior work~\cite{vsf-fabric} is VisuoSpatial Foresight (VSF), an extension of Visual Foresight~\cite{visual_foresight_2018}. It consists of a training phase followed by a planning phase. An overview of \vsf is provided in Section~\ref{sec:approach}, and practical implementation details are in Section~\ref{sec:implementation}. For the planning phase described in Section~\ref{ssec:planning}, we tuned the hyperparameters in Table~\ref{tab:vismpc-hyperparams}. The CEM variance reported is the diagonal covariance used for folding and double folding. We found that for smoothing, a lower CEM variance (0.25, 0.25, 0.04, 0.04) results in better performance, though it may encourage the policy towards taking shorter actions. For CMA-ES, we use the open source Python implementation PyCMA (\url{https://pypi.org/project/cma/}), changing only the number of iterations, initial mean, and initial variance from default parameters. CMA-ES and CEM take a similar amount of computation time.

As described in Section~\ref{ssec:costs}, we evaluate with a Pixel L2 and learned Vertex L2 cost function. For the Pixel L2 cost function (Equation~\ref{eq:cost}), we remove the 7 pixels on each side of the image to get rid of the impact of the dark border, using only the inner $42\times 42$ region of the $56\times 56$ image. For the Vertex L2 cost function, as described in Section~\ref{sssec:cost2}, we generate a second dataset from the primary dataset collected. Each of the 9,932 episodes in \dnew can contribute up to $10 \choose 2$ image pairs to use in the second dataset, but we sample only 10 of these possible pairs from each episode to keep the dataset size modest (and the same size as the primary dataset). Specifically, we use the following 10 pairs, chosen for their variable gaps in temporal distance: \{$(\bo_1, \bo_2)$, $(\bo_1, \bo_3)$, $(\bo_1, \bo_5)$, $(\bo_1, \bo_9)$,  $(\bo_6, \bo_8)$, $(\bo_6, \bo_{10})$, $(\bo_6, \bo_7)$, $(\bo_3, \bo_4)$, $(\bo_3, \bo_7)$, $(\bo_3, \bo_9) $\}. During training, we flip the order of half of the data points to encourage the network to ignore the direction of time in its estimation of mesh distance. As mentioned in Section~\ref{sssec:cost2}, we annotate all data points with the sum of the distances between corresponding points in the ground truth mesh states, i.e. $$\sum_{i=0}^{625} ||p_1^{(i)} - p_2^{(i)} ||_2^2 $$ where $p_1^{(i)}$ is the $(x,y,z)$ coordinates of the $i$-th point of the mesh shown in the first image and $p_2^{(i)}$ is the $(x,y,z)$ coordinates of the $i$-th point of the mesh in the second image. We divide all labels by the maximum value for more stable training. Finally, for the network architecture, we use the same CNN as the actor in the DAgger baseline as described in Section~\ref{app:dagger}. However, to accommodate the second image input, we pass both images through the same convolutional layers and concatenate the outputs to a 5184-dimensional vector before applying the fully connected layers. The resulting network has about 1.5 million parameters.

\section{Supplementary Smoothing Results}\label{app:results}

\subsection{Statistical Significance Tests}

We run the Mann-Whitney U test~\cite{mannwhitney} on the coverage and number of action results reported in Table~\ref{tab:analytic} for \vsfrss against all baselines other than Imitation Learning, to which we wish to perform similarly. See Table~\ref{tab:stat} for computed $p$-values. We conclude that we can confidently reject the null hypothesis that the values are drawn from the same distribution for all metrics except Tier 2 coverage for Wrinkle and the Tier 1 and Tier 3 number of actions for DDPG ($p$ $<$ 0.02). Note that Tier 3 results are most informative, as it is the most difficult tier.

\begin{table}[t]
\caption{
Mann-Whitney Test $p$-values for coverage and number of actions of \vsfrss compared with Random, Highest, Wrinkle and DDPG baselines across all tiers of difficulty for smoothing.
}
\centering
\begin{tabular}{c c c c}
\textbf{Tier} & \textbf{Policy} & \textbf{Coverage $p$-value} & \textbf{Actions $p$-value} \\  \hline
1 & Random & 0.0000 & 0.0000 \\
1 & Highest & 0.0000 & 0.0002 \\
1 & Wrinkle & 0.0040 & 0.0015 \\
1 & DDPG & 0.0044 & 0.3670 \\
2 & Random & 0.0000 & 0.0000 \\
2 & Highest & 0.0000 & 0.0000 \\
2 & Wrinkle & 0.2323 & 0.0091 \\
2 & DDPG & 0.0000 & 0.0000 \\
3 & Random & 0.0000 & 0.0000 \\
3 & Highest & 0.0000 & 0.0000 \\
3 & Wrinkle & 0.0030 & 0.0199 \\
3 & DDPG & 0.0000 & 0.0674 \\
\end{tabular}
\label{tab:stat}
\end{table}

\subsection{Domain Randomization Ablation}

For \dold, we run 50 simulated smoothing episodes per tier with a policy trained \textit{without} domain randomization and compare with the 200 episodes from Table~\ref{tab:analytic}. In the episodes without domain randomization, we keep fabric color, camera angle, background plane shading, and brightness constant at training and testing time. In the episodes with domain randomization, we randomize these parameters in the training data and test in the same setting as the experiments without domain randomization, which can be interpreted as a random initialization of the domain randomized parameters. In particular, we vary the following:

\begin{itemize}
    \item Fabric color RGB values uniformly between (0, 0, 128) and (115, 179, 255), centered around blue.
    \item Background plane color RGB values uniformly between (102, 102, 102) and (153, 153, 153).
    \item RGB gamma correction with gamma uniformly between 0.7 and 1.3.
    \item A fixed amount to subtract from the depth image between 40 and 50 to simulate changing the height of the depth camera.
    \item Camera position ($x, y, z$) as $(0.5+\delta_1, 0.5+\delta_2, 1.45+\delta_3)$ meters, where each $\delta_i$ is sampled from $\mathcal{N}(0, 0.04)$.
    \item Camera rotation with Euler angles sampled from $\mathcal{N}(0, 90^{\degree} )$.
    \item Random noise at each pixel uniformly between -15 and 15.
\end{itemize}

From the results in Table~\ref{tab:dr}, we find that final coverage values are similar whether or not we use domain randomization on training data, suggesting our domain randomization techniques do not have an adverse effect on performance in simulation.

To analyze robustness of the policy to variation in the randomized parameters, we also evaluate the former two policies (trained with and without domain randomization) with randomization in the test environment on Tier 3 starting states. Specifically, we change the color of the fabric in fixed increments from its non-randomized setting (RGB (25, 89, 217)) until performance starts to deteriorate. In Table~\ref{tab:dr2}, we observe that the domain randomized policy maintains high coverage within the training range (RGB (0, 0, 128) to (115, 179, 255)) while the policy without domain randomization suffers as soon as the fabric color is slightly altered.

\begin{table}[t]
\caption{
Coverage and number of actions for simulated smoothing episodes from \dold, with and without domain randomization on training data, where the domain randomized results are from Table~\ref{tab:analytic}.
}
\centering
\begin{tabular}{c c r r}
\textbf{Tier} & \textbf{Domain Randomized?} & \textbf{Coverage} & \textbf{Actions} \\ \hline
1 & Yes & 92.5 $\pm$ \:\:2.5 & 8.3 $\pm$ 4.7 \\
1 & No & 93.0 $\pm$ \:\:3.0 & 6.9 $\pm$ 4.1 \\
2 & Yes & 90.3 $\pm$ \:\:3.8 & 12.1 $\pm$ 3.4 \\ 
2 & No & 91.2 $\pm$ \:\:9.2 & 8.7 $\pm$ 3.6 \\
3 & Yes & 89.3 $\pm$ \:\:5.9 & 13.1 $\pm$ 2.9 \\
3 & No & 85.1 $\pm$ 12.8 & 9.9 $\pm$ 3.9 \\
\end{tabular}
\label{tab:dr}
\end{table}

\begin{table}[t]
\caption{
Coverage and number of actions for Tier 3 simulated smoothing episodes with and without domain randomization on \dold training data, where we vary fabric color in fixed increments. (26, 89, 217) is the default blue color and (128, 191, 115) is slightly outside the domain randomization range. Values for the default setting are repeated from Table~\ref{tab:dr} and all other data points are averaged over 20 episodes.
}
\centering
\begin{tabular}{c c r r}
\textbf{RGB Values} & \textbf{DR?} & \textbf{Coverage} & \textbf{Actions} \\ \hline
(26, 89, 217) & Yes & 89.3 $\pm$ \:\:5.9 & 13.1 $\pm$ 2.9 \\
(51, 115, 191) & Yes & 89.3 $\pm$ 10.3 & 11.7 $\pm$ 3.5 \\
(77, 140, 166) & Yes & 91.4 $\pm$ \:\:3.1 & 11.7 $\pm$ 3.1 \\ 
(102, 165, 140) & Yes & 85.6 $\pm$ 10.1 & 13.2 $\pm$ 2.7 \\
(128, 191, 115) & Yes & 54.7 $\pm$ \:\:6.5 & 10.3 $\pm$ 4.0 \\
(26, 89, 217) & No & 85.1 $\pm$ 12.8 & 9.9 $\pm$ 3.9 \\
(51, 115, 191) & No & 60.7 $\pm$ 13.6 & 7.4 $\pm$ 2.4 \\
\end{tabular}
\label{tab:dr2}
\end{table}




%% file: 0-main.bbl
\begin{thebibliography}{87}
\providecommand{\natexlab}[1]{#1}
\providecommand{\url}[1]{{#1}}
\providecommand{\urlprefix}{URL }
\expandafter\ifx\csname urlstyle\endcsname\relax
  \providecommand{\doi}[1]{DOI~\discretionary{}{}{}#1}\else
  \providecommand{\doi}{DOI~\discretionary{}{}{}\begingroup
  \urlstyle{rm}\Url}\fi
\providecommand{\eprint}[2][]{\url{#2}}

\bibitem[{Babaeizadeh et~al.(2018)Babaeizadeh, Finn, Erhan, Campbell, and
  Levine}]{sv2p}
Babaeizadeh M, Finn C, Erhan D, Campbell RH, Levine S (2018) {Stochastic
  Variational Video Prediction}. In: International Conference on Learning
  Representations (ICLR)

\bibitem[{Balaguer and Carpin(2011)}]{balaguer2011combining}
Balaguer B, Carpin S (2011) {Combining Imitation and Reinforcement Learning to
  Fold Deformable Planar Objects}. In: IEEE/RSJ International Conference on
  Intelligent Robots and Systems (IROS)

\bibitem[{Balakrishna et~al.(2019)Balakrishna, Thananjeyan, Lee, Zahed, Li,
  Gonzalez, and Goldberg}]{converging-supervisor}
Balakrishna A, Thananjeyan B, Lee J, Zahed A, Li F, Gonzalez JE, Goldberg K
  (2019) {On-Policy Robot Imitation Learning from a Converging Supervisor}. In:
  Conference on Robot Learning (CoRL)

\bibitem[{Baraff and Witkin(1998)}]{cloth-cloth-collisions}
Baraff D, Witkin A (1998) {Large Steps in Cloth Simulation}. In: ACM SIGGRAPH

\bibitem[{Berkenkamp et~al.(2016)Berkenkamp, Schoellig, and
  Krause}]{berkenkamp2016safe}
Berkenkamp F, Schoellig AP, Krause A (2016) {Safe Controller Optimization for
  Quadrotors with Gaussian Processes}. In: IEEE International Conference on
  Robotics and Automation (ICRA)

\bibitem[{Borras et~al.(2019)Borras, Alenya, and
  Torras}]{grasp_centered_survey_2019}
Borras J, Alenya G, Torras C (2019) {A Grasping-centered Analysis for Cloth
  Manipulation}. arXiv preprint arXiv:190608202

\bibitem[{Chen et~al.(2019)Chen, Zhou, Koltun, and Krahenbuhl}]{cheating_2019}
Chen D, Zhou B, Koltun V, Krahenbuhl P (2019) {Learning by Cheating}. In:
  Conference on Robot Learning (CoRL)

\bibitem[{Chiuso and Pillonetto(2019)}]{system-id}
Chiuso A, Pillonetto G (2019) {System Identification: A Machine Learning
  Perspective}. Annual Review of Control, Robotics, and Autonomous Systems

\bibitem[{Chua et~al.(2018)Chua, Calandra, McAllister, and
  Levine}]{handful-of-trials}
Chua K, Calandra R, McAllister R, Levine S (2018) {Deep Reinforcement Learning
  in a Handful of Trials Using Probabilistic Dynamics Models}. In: Neural
  Information Processing Systems (NeurIPS)

\bibitem[{Community(2018)}]{blender}
Community BO (2018) Blender - a 3D modelling and rendering package. Blender
  Foundation, Stichting Blender Foundation, Amsterdam,
  \urlprefix\url{http://www.blender.org}

\bibitem[{Coumans and Bai(2016--2019)}]{coumans2019}
Coumans E, Bai Y (2016--2019) Pybullet, a python module for physics simulation
  for games, robotics and machine learning. \url{http://pybullet.org}

\bibitem[{Dasari et~al.(2019)Dasari, Ebert, Tian, Nair, Bucher, Schmeckpeper,
  Singh, Levine, and Finn}]{robonet}
Dasari S, Ebert F, Tian S, Nair S, Bucher B, Schmeckpeper K, Singh S, Levine S,
  Finn C (2019) {RoboNet: Large-Scale Multi-Robot Learning}. In: Conference on
  Robot Learning (CoRL)

\bibitem[{Denton and Fergus(2018)}]{svg}
Denton E, Fergus R (2018) {Stochastic Video Generation with a Learned Prior}.
  In: International Conference on Machine Learning (ICML)

\bibitem[{Dhariwal et~al.(2017)Dhariwal, Hesse, Klimov, Nichol, Plappert,
  Radford, Schulman, Sidor, Wu, and Zhokhov}]{baselines}
Dhariwal P, Hesse C, Klimov O, Nichol A, Plappert M, Radford A, Schulman J,
  Sidor S, Wu Y, Zhokhov P (2017) {OpenAI Baselines}.
  \url{https://github.com/openai/baselines}

\bibitem[{Doumanoglou et~al.(2014)Doumanoglou, Kargakos, Kim, and
  Malassiotis}]{unfolding_rf_2014}
Doumanoglou A, Kargakos A, Kim TK, Malassiotis S (2014) {Autonomous Active
  Recognition and Unfolding of Clothes Using Random Decision Forests and
  Probabilistic Planning}. In: IEEE International Conference on Robotics and
  Automation (ICRA)

\bibitem[{Ebert et~al.(2017)Ebert, Finn, Lee, and Levine}]{bair_push_2017}
Ebert F, Finn C, Lee AX, Levine S (2017) {Self-Supervised Visual Planning with
  Temporal Skip Connections}. In: Conference on Robot Learning (CoRL)

\bibitem[{Ebert et~al.(2018)Ebert, Finn, Dasari, Xie, Lee, and
  Levine}]{visual_foresight_2018}
Ebert F, Finn C, Dasari S, Xie A, Lee A, Levine S (2018) {Visual Foresight:
  Model-Based Deep Reinforcement Learning for Vision-Based Robotic Control}.
  arXiv preprint arXiv:181200568

\bibitem[{Erickson et~al.(2018)Erickson, Clever, Turk, Liu, and
  Kemp}]{deep_dressing_2018}
Erickson Z, Clever HM, Turk G, Liu CK, Kemp CC (2018) {Deep Haptic Model
  Predictive Control for Robot-Assisted Dressing}. In: IEEE International
  Conference on Robotics and Automation (ICRA)

\bibitem[{Erickson et~al.(2020)Erickson, Gangaram, Kapusta, Liu, and
  Kemp}]{erickson2020assistivegym}
Erickson Z, Gangaram V, Kapusta A, Liu CK, Kemp CC (2020) {Assistive Gym: A
  Physics Simulation Framework for Assistive Robotics}. In: IEEE International
  Conference on Robotics and Automation (ICRA)

\bibitem[{Finn and Levine(2017)}]{finn_vf_2017}
Finn C, Levine S (2017) {Deep Visual Foresight for Planning Robot Motion}. In:
  IEEE International Conference on Robotics and Automation (ICRA)

\bibitem[{Finn et~al.(2016)Finn, Goodfellow, and Levine}]{finn_cdna}
Finn C, Goodfellow I, Levine S (2016) {Unsupervised Learning for Physical
  Interaction through Video Prediction}. In: Neural Information Processing
  Systems (NeurIPS)

\bibitem[{Ganapathi et~al.(2020)Ganapathi, Sundaresan, Thananjeyan,
  Balakrishna, Seita, Hoque, Gonzalez, and Goldberg}]{mmgsd}
Ganapathi A, Sundaresan P, Thananjeyan B, Balakrishna A, Seita D, Hoque R,
  Gonzalez JE, Goldberg K (2020) Mmgsd: Multi-modal gaussian shape descriptors
  for correspondence matching in 1d and 2d deformable objects. In:
  International Conference on Intelligent Robots and Systems (IROS) Workshop on
  Managing Deformation, IEEE

\bibitem[{Ganapathi et~al.(2021)Ganapathi, Sundaresan, Thananjeyan,
  Balakrishna, Seita, Grannen, Hwang, Hoque, Gonzalez, Jamali, Yamane, Iba, and
  Goldberg}]{ganapathi2020learning}
Ganapathi A, Sundaresan P, Thananjeyan B, Balakrishna A, Seita D, Grannen J,
  Hwang M, Hoque R, Gonzalez JE, Jamali N, Yamane K, Iba S, Goldberg K (2021)
  {Learning Dense Visual Correspondences in Simulation to Smooth and Fold Real
  Fabrics}. In: IEEE International Conference on Robotics and Automation (ICRA)

\bibitem[{Gao et~al.(2016)Gao, Chang, and Demiris}]{personalized_dressing_2016}
Gao Y, Chang HJ, Demiris Y (2016) {Iterative Path Optimisation for Personalised
  Dressing Assistance using Vision and Force Information}. In: IEEE/RSJ
  International Conference on Intelligent Robots and Systems (IROS)

\bibitem[{Goodfellow et~al.(2016)Goodfellow, Bengio, and
  Courville}]{Goodfellow-et-al-2016}
Goodfellow I, Bengio Y, Courville A (2016) {Deep Learning}. MIT Press,
  \url{http://www.deeplearningbook.org}

\bibitem[{Hansen and Auger(2011)}]{CMA-ES}
Hansen N, Auger A (2011) {CMA-ES: Evolution Strategies and Covariance Matrix
  Adaptation}. Association for Computing Machinery, New York, NY, USA

\bibitem[{Hewing et~al.(2018)Hewing, Liniger, and Zeilinger}]{cautious-MPC}
Hewing L, Liniger A, Zeilinger M (2018) {Cautious NMPC with Gaussian Process
  Dynamics for Autonomous Miniature Race Cars}. In: European Controls
  Conference (ECC)

\bibitem[{Hochreiter and Schmidhuber(1997)}]{hochreiter1997long}
Hochreiter S, Schmidhuber J (1997) {Long Short-Term Memory}. Neural Computation
  9

\bibitem[{Hoque et~al.(2020)Hoque, Seita, Balakrishna, Ganapathi, Tanwani,
  Jamali, Yamane, Iba, and Goldberg}]{vsf-fabric}
Hoque R, Seita D, Balakrishna A, Ganapathi A, Tanwani AK, Jamali N, Yamane K,
  Iba S, Goldberg K (2020) {VisuoSpatial Foresight for Multi-Step, Multi-Task
  Fabric Manipulation}. In: Robotics: Science and Systems (RSS)

\bibitem[{Jangir et~al.(2020)Jangir, Alenya, and Torras}]{rishabh_2019}
Jangir R, Alenya G, Torras C (2020) {Dynamic Cloth Manipulation with Deep
  Reinforcement Learning}. In: IEEE International Conference on Robotics and
  Automation (ICRA)

\bibitem[{Jia et~al.(2018)Jia, Hu, Pan, and Manocha}]{jia_visual_feedback_2018}
Jia B, Hu Z, Pan J, Manocha D (2018) {Manipulating Highly Deformable Materials
  Using a Visual Feedback Dictionary}. In: IEEE International Conference on
  Robotics and Automation (ICRA)

\bibitem[{Jia et~al.(2019)Jia, Pan, Hu, Pan, and
  Manocha}]{jia_cloth_manip_2019}
Jia B, Pan Z, Hu Z, Pan J, Manocha D (2019) {Cloth Manipulation Using
  Random-Forest-Based Imitation Learning}. In: IEEE International Conference on
  Robotics and Automation (ICRA)

\bibitem[{Kazanzides et~al.(2014)Kazanzides, Chen, Deguet, Fischer, Taylor, and
  DiMaio}]{dvrk2014}
Kazanzides P, Chen Z, Deguet A, Fischer G, Taylor R, DiMaio S (2014) {An
  Open-Source Research Kit for the da Vinci Surgical System}. In: IEEE
  International Conference on Robotics and Automation (ICRA)

\bibitem[{Kingma and Ba(2015)}]{adam2015}
Kingma DP, Ba J (2015) {Adam: A Method for Stochastic Optimization}. In:
  International Conference on Learning Representations (ICLR)

\bibitem[{Kita et~al.(2009{\natexlab{a}})Kita, Ueshiba, Neo, and
  Kita}]{kita_2009_iros}
Kita Y, Ueshiba T, Neo ES, Kita N (2009{\natexlab{a}}) {A Method For Handling a
  Specific Part of Clothing by Dual Arms}. In: IEEE/RSJ International
  Conference on Intelligent Robots and Systems (IROS)

\bibitem[{Kita et~al.(2009{\natexlab{b}})Kita, Ueshiba, Neo, and
  Kita}]{kita_2009_icra}
Kita Y, Ueshiba T, Neo ES, Kita N (2009{\natexlab{b}}) {Clothes State
  Recognition Using 3D Observed Data}. In: IEEE International Conference on
  Robotics and Automation (ICRA)

\bibitem[{Kocijan et~al.(2004)Kocijan, Murray-Smith, Rasmussen, and
  Girard}]{GP-MPC}
Kocijan J, Murray-Smith R, Rasmussen C, Girard A (2004) {Gaussian Process Model
  Based Predictive Control}. In: American Control Conference (ACC)

\bibitem[{Kumar et~al.(2020)Kumar, Babaeizadeh, Erhan, Finn, Levine, Dinh, and
  Kingma}]{videoflow}
Kumar M, Babaeizadeh M, Erhan D, Finn C, Levine S, Dinh L, Kingma D (2020)
  {VideoFlow: A Conditional Flow-Based Model for Stochastic Video Generation}.
  In: International Conference on Learning Representations (ICLR)

\bibitem[{Lee et~al.(2018)Lee, Zhang, Ebert, Abbeel, Finn, and Levine}]{savp}
Lee AX, Zhang R, Ebert F, Abbeel P, Finn C, Levine S (2018) {Stochastic
  Adversarial Video Prediction}. arXiv preprint arXiv:180401523

\bibitem[{Lee et~al.(2020)Lee, Ward, Cosgun, Dasagi, Corke, and
  Leitner}]{lee2020learning}
Lee R, Ward D, Cosgun A, Dasagi V, Corke P, Leitner J (2020) {Learning
  Arbitrary-Goal Fabric Folding with One Hour of Real Robot Experience}. In:
  Conference on Robot Learning (CoRL)

\bibitem[{Li et~al.(2015)Li, Yue, Grinspun, and Allen}]{folding_iros_2015}
Li Y, Yue Y, Grinspun DXE, Allen PK (2015) {Folding Deformable Objects using
  Predictive Simulation and Trajectory Optimization}. In: IEEE/RSJ
  International Conference on Intelligent Robots and Systems (IROS)

\bibitem[{Li et~al.(2016)Li, Hu, Xu, Yue, Grinspun, and Allen}]{ironing_2016}
Li Y, Hu X, Xu D, Yue Y, Grinspun E, Allen PK (2016) {Multi-Sensor Surface
  Analysis for Robotic Ironing}. In: IEEE International Conference on Robotics
  and Automation (ICRA)

\bibitem[{Lillicrap et~al.(2016)Lillicrap, Hunt, Pritzel, Heess, Erez, Tassa,
  Silver, and Wierstra}]{ddpg2016}
Lillicrap TP, Hunt JJ, Pritzel A, Heess N, Erez T, Tassa Y, Silver D, Wierstra
  D (2016) {Continuous Control with Deep Reinforcement Learning}. In:
  International Conference on Learning Representations (ICLR)

\bibitem[{Lin et~al.(2020)Lin, Wang, Olkin, and Held}]{lin2020softgym}
Lin X, Wang Y, Olkin J, Held D (2020) {SoftGym: Benchmarking Deep Reinforcement
  Learning for Deformable Object Manipulation}. In: Conference on Robot
  Learning (CoRL)

\bibitem[{Lippi et~al.(2020)Lippi, Poklukar, Welle, Varava, Yin, Marino, and
  Kragic}]{latentroadmap}
Lippi M, Poklukar P, Welle MC, Varava A, Yin H, Marino A, Kragic D (2020)
  {Latent Space Roadmap for Visual Action Planning of Deformable and Rigid
  Object Manipulation}. In: IEEE/RSJ International Conference on Intelligent
  Robots and Systems (IROS)

\bibitem[{Maitin-Shepard et~al.(2010)Maitin-Shepard, Cusumano-Towner, Lei, and
  Abbeel}]{maitin2010cloth}
Maitin-Shepard J, Cusumano-Towner M, Lei J, Abbeel P (2010) {Cloth Grasp Point
  Detection Based on Multiple-View Geometric Cues with Application to Robotic
  Towel Folding}. In: IEEE International Conference on Robotics and Automation
  (ICRA)

\bibitem[{Mann and Whitney(1947)}]{mannwhitney}
Mann H, Whitney D (1947) {On a Test of Whether One of Two Random Variables is
  Stochastically Larger than the Other}. Annals of Mathematical Statistics

\bibitem[{Matas et~al.(2018)Matas, James, and Davison}]{sim2real_deform_2018}
Matas J, James S, Davison AJ (2018) {Sim-to-Real Reinforcement Learning for
  Deformable Object Manipulation}. Conference on Robot Learning (CoRL)

\bibitem[{Miller et~al.(2012)Miller, van~den Berg, Fritz, Darrell, Goldberg,
  and Abbeel}]{laundry2012}
Miller S, van~den Berg J, Fritz M, Darrell T, Goldberg K, Abbeel P (2012) {A
  Geometric Approach to Robotic Laundry Folding}. In: International Journal of
  Robotics Research (IJRR)

\bibitem[{Nagabandi et~al.(2018)Nagabandi, Kahn, Fearing, and
  Levine}]{nagabandi_2018}
Nagabandi A, Kahn G, Fearing R, Levine S (2018) {Neural Network Dynamics for
  Model-Based Deep Reinforcement Learning with Model-Free Fine-Tuning}. In:
  IEEE International Conference on Robotics and Automation (ICRA)

\bibitem[{Nair et~al.(2018)Nair, McGrew, Andrychowicz, Zaremba, and
  Abbeel}]{overcoming_exploration}
Nair A, McGrew B, Andrychowicz M, Zaremba W, Abbeel P (2018) {Overcoming
  Exploration in Reinforcement Learning with Demonstrations}. In: IEEE
  International Conference on Robotics and Automation (ICRA)

\bibitem[{Nair and Finn(2020)}]{gap_suraj_2020}
Nair S, Finn C (2020) {Goal-Aware Prediction: Learning to Model What Matters}.
  In: International Conference on Machine Learning (ICML)

\bibitem[{Nair et~al.(2020)Nair, Babaeizadeh, Finn, Levine, and
  Kumar}]{time_reversal}
Nair S, Babaeizadeh M, Finn C, Levine S, Kumar V (2020) {Time Reversal as
  Self-Supervision}. In: IEEE International Conference on Robotics and
  Automation (ICRA)

\bibitem[{Narain et~al.(2012)Narain, Samii, and O'Brien}]{arcsim2012}
Narain R, Samii A, O'Brien JF (2012) {Adaptive Anisotropic Remeshing for Cloth
  Simulation}. In: ACM SIGGRAPH Asia

\bibitem[{OpenAI et~al.(2019)OpenAI, Andrychowicz, Baker, Chociej, Jozefowicz,
  McGrew, Pachocki, Petron, Plappert, Powell, Ray, Schneider, Sidor, Tobin,
  Welinder, Weng, and Zaremba}]{openai-dactyl}
OpenAI, Andrychowicz M, Baker B, Chociej M, Jozefowicz R, McGrew B, Pachocki J,
  Petron A, Plappert M, Powell G, Ray A, Schneider J, Sidor S, Tobin J,
  Welinder P, Weng L, Zaremba W (2019) {Learning Dexterous In-Hand
  Manipulation}. In: International Journal of Robotics Research (IJRR)

\bibitem[{Osawa et~al.(2007)Osawa, Seki, and Kamiya}]{osawa_2007}
Osawa F, Seki H, Kamiya Y (2007) {Unfolding of Massive Laundry and
  Classification Types by Dual Manipulator}. Journal of Advanced Computational
  Intelligence and Intelligent Informatics 11(5)

\bibitem[{Pomerleau(1991)}]{Pomerleau_behavior_cloning}
Pomerleau DA (1991) {Efficient Training of Artificial Neural Networks for
  Autonomous Navigation}. Neural Comput 3

\bibitem[{Provot(1995)}]{provot_1996}
Provot X (1995) {Deformation Constraints in a Mass-Spring Model to Describe
  Rigid Cloth Behavior}. In: Graphics Interface

\bibitem[{Radford et~al.(2016)Radford, Metz, and Chintala}]{dcgan2016}
Radford A, Metz L, Chintala S (2016) {Unsupervised Representation Learning with
  Deep Convolutional Generative Adversarial Networks}. In: International
  Conference on Learning Representations (ICLR)

\bibitem[{Rosolia and Borrelli(2019)}]{MPCRacing}
Rosolia U, Borrelli F (2019) {Learning how to Autonomously Race a Car: a
  Predictive Control Approach}. In: IEEE Transactions on Control Systems
  Technology

\bibitem[{Ross et~al.(2011)Ross, Gordon, and Bagnell}]{ross2011reduction}
Ross S, Gordon GJ, Bagnell JA (2011) {A Reduction of Imitation Learning and
  Structured Prediction to No-Regret Online Learning}. In: International
  Conference on Artificial Intelligence and Statistics (AISTATS)

\bibitem[{Rubinstein(1999)}]{cem_1999}
Rubinstein R (1999) {The Cross-Entropy Method for Combinatorial and Continuous
  Optimization}. Methodology And Computing In Applied Probability

\bibitem[{Sanchez et~al.(2018)Sanchez, Corrales, Bouzgarrou, and
  Mezouar}]{manip_deformable_survey_2018}
Sanchez J, Corrales JA, Bouzgarrou BC, Mezouar Y (2018) {Robotic Manipulation
  and Sensing of Deformable Objects in Domestic and Industrial Applications: a
  Survey}. In: International Journal of Robotics Research (IJRR)

\bibitem[{Schrimpf and Wetterwald(2012)}]{sewing_2012}
Schrimpf J, Wetterwald LE (2012) {Experiments Towards Automated Sewing With a
  Multi-Robot System}. In: IEEE International Conference on Robotics and
  Automation (ICRA)

\bibitem[{Seita et~al.(2018)Seita, Krishnan, Fox, McKinley, Canny, and
  Goldberg}]{seita_icra_2018}
Seita D, Krishnan S, Fox R, McKinley S, Canny J, Goldberg K (2018) {Fast and
  Reliable Autonomous Surgical Debridement with Cable-Driven Robots Using a
  Two-Phase Calibration Procedure}. In: IEEE International Conference on
  Robotics and Automation (ICRA)

\bibitem[{Seita et~al.(2019)Seita, Jamali, Laskey, Berenstein, Tanwani,
  Baskaran, Iba, Canny, and Goldberg}]{seita-bedmaking}
Seita D, Jamali N, Laskey M, Berenstein R, Tanwani AK, Baskaran P, Iba S, Canny
  J, Goldberg K (2019) {Deep Transfer Learning of Pick Points on Fabric for
  Robot Bed-Making}. In: International Symposium on Robotics Research (ISRR)

\bibitem[{Seita et~al.(2020)Seita, Ganapathi, Hoque, Hwang, Cen, Tanwani,
  Balakrishna, Thananjeyan, Ichnowski, Jamali, Yamane, Iba, Canny, and
  Goldberg}]{seita_ryan}
Seita D, Ganapathi A, Hoque R, Hwang M, Cen E, Tanwani AK, Balakrishna A,
  Thananjeyan B, Ichnowski J, Jamali N, Yamane K, Iba S, Canny J, Goldberg K
  (2020) {Deep Imitation Learning of Sequential Fabric Smoothing From an
  Algorithmic Supervisor}. In: IEEE/RSJ International Conference on Intelligent
  Robots and Systems (IROS)

\bibitem[{Seita et~al.(2021)Seita, Florence, Tompson, Coumans, Sindhwani,
  Goldberg, and Zeng}]{seita2020learning}
Seita D, Florence P, Tompson J, Coumans E, Sindhwani V, Goldberg K, Zeng A
  (2021) {Learning to Rearrange Deformable Cables, Fabrics, and Bags with
  Goal-Conditioned Transporter Networks}. In: IEEE International Conference on
  Robotics and Automation (ICRA)

\bibitem[{Shibata et~al.(2012)Shibata, Yoshimi, Mizukawa, and
  Ando}]{shibata2012trajectory}
Shibata S, Yoshimi T, Mizukawa M, Ando Y (2012) {A Trajectory Generation of
  Cloth Object Folding Motion Toward Realization of Housekeeping Robot}. In:
  International Conference on Ubiquitous Robots and Ambient Intelligence (URAI)

\bibitem[{Shin et~al.(2019)Shin, Ferguson, Pedram, Ma, Dutson, and
  Rosen}]{rosen_icra_tissues_2019}
Shin C, Ferguson PW, Pedram SA, Ma J, Dutson EP, Rosen J (2019) {Autonomous
  Tissue Manipulation via Surgical Robot Using Learning Based Model Predictive
  Control}. In: IEEE International Conference on Robotics and Automation (ICRA)

\bibitem[{Sun et~al.(2014)Sun, Aragon-Camarasa, Cockshott, Rogers, and
  Siebert}]{heuristic_wrinkles_2014}
Sun L, Aragon-Camarasa G, Cockshott P, Rogers S, Siebert JP (2014) {A
  Heuristic-Based Approach for Flattening Wrinkled Clothes}. Towards Autonomous
  Robotic Systems TAROS 2013 Lecture Notes in Computer Science, vol 8069

\bibitem[{Sun et~al.(2015)Sun, Aragon-Camarasa, Rogers, and
  Siebert}]{cloth_icra_2015}
Sun L, Aragon-Camarasa G, Rogers S, Siebert JP (2015) {Accurate Garment Surface
  Analysis using an Active Stereo Robot Head with Application to Dual-Arm
  Flattening}. In: IEEE International Conference on Robotics and Automation
  (ICRA)

\bibitem[{Thananjeyan et~al.(2017)Thananjeyan, Garg, Krishnan, Chen, Miller,
  and Goldberg}]{thananjeyan2017multilateral}
Thananjeyan B, Garg A, Krishnan S, Chen C, Miller L, Goldberg K (2017)
  {Multilateral Surgical Pattern Cutting in 2D Orthotropic Gauze with Deep
  Reinforcement Learning Policies for Tensioning}. In: IEEE International
  Conference on Robotics and Automation (ICRA)

\bibitem[{Thananjeyan* et~al.(2020)Thananjeyan*, Balakrishna*, Nair, Luo,
  Srinivasan, Hwang, Gonzalez, Ibarz, Finn, and Goldberg}]{recovery-rl}
Thananjeyan* B, Balakrishna* A, Nair S, Luo M, Srinivasan K, Hwang M, Gonzalez
  JE, Ibarz J, Finn C, Goldberg K (2020) {Recovery RL: Safe Reinforcement
  Learning with Learned Recovery Zones}. In: NeurIPS Robot Learning Workshop,
  NeurIPS

\bibitem[{Thananjeyan et~al.(2020)Thananjeyan, Balakrishna, Rosolia, Li,
  McAllister, Gonzalez, Levine, Borrelli, and Goldberg}]{thananjeyan2019safety}
Thananjeyan B, Balakrishna A, Rosolia U, Li F, McAllister R, Gonzalez JE,
  Levine S, Borrelli F, Goldberg K (2020) {Safety Augmented Value Estimation
  from Demonstrations (SAVED): Safe Deep Model-Based RL for Sparse Cost Robotic
  Tasks}. In: IEEE Robotics and Automation Letters (RA-L)

\bibitem[{Tobin et~al.(2017)Tobin, Fong, Ray, Schneider, Zaremba, and
  Abbeel}]{domain_randomization}
Tobin J, Fong R, Ray A, Schneider J, Zaremba W, Abbeel P (2017) {Domain
  Randomization for Transferring Deep Neural Networks from Simulation to the
  Real World}. In: IEEE/RSJ International Conference on Intelligent Robots and
  Systems (IROS)

\bibitem[{Todorov et~al.(2012)Todorov, Erez, and Tassa}]{mujoco}
Todorov E, Erez T, Tassa Y (2012) {MuJoCo: A Physics Engine for Model-Based
  Control}. In: IEEE/RSJ International Conference on Intelligent Robots and
  Systems (IROS)

\bibitem[{Torgerson and Paul(1987)}]{Torgerson1987VisionGR}
Torgerson E, Paul F (1987) {Vision Guided Robotic Fabric Manipulation for
  Apparel Manufacturing}. In: IEEE International Conference on Robotics and
  Automation (ICRA)

\bibitem[{Vaswani et~al.(2018)Vaswani, Bengio, Brevdo, Chollet, Gomez, Gouws,
  Jones, Kaiser, Kalchbrenner, Parmar, Sepassi, Shazeer, and
  Uszkoreit}]{tensor2tensor}
Vaswani A, Bengio S, Brevdo E, Chollet F, Gomez AN, Gouws S, Jones L, Kaiser L,
  Kalchbrenner N, Parmar N, Sepassi R, Shazeer N, Uszkoreit J (2018)
  Tensor2tensor for neural machine translation. CoRR abs/1803.07416,
  \urlprefix\url{http://arxiv.org/abs/1803.07416}

\bibitem[{Vecerik et~al.(2017)Vecerik, Hester, Scholz, Wang, Pietquin, Piot,
  Heess, Rothörl, Lampe, and Riedmiller}]{ddpgfd}
Vecerik M, Hester T, Scholz J, Wang F, Pietquin O, Piot B, Heess N, Rothörl T,
  Lampe T, Riedmiller M (2017) {Leveraging Demonstrations for Deep
  Reinforcement Learning on Robotics Problems with Sparse Rewards}. arXiv
  preprint arXiv:170708817

\bibitem[{Verlet(1967)}]{verlet_1967}
Verlet L (1967) {Computer Experiments on Classical Fluids: I. Theormodynamical
  Properties of Lennard-Jones Molecules}. Physics Review 159(98)

\bibitem[{Wang et~al.(2004)Wang, Bovik, Sheikh, and Simoncelli}]{ssim_2004}
Wang Z, Bovik AC, Sheikh HR, Simoncelli EP (2004) {Image Quality Assessment:
  From Error Visibility to Structural Similarity}. Trans Img Proc

\bibitem[{Willimon et~al.(2011)Willimon, Birchfield, and
  Walker}]{willimon_unfolding_laundry_2011}
Willimon B, Birchfield S, Walker I (2011) {Model for Unfolding Laundry using
  Interactive Perception}. In: IEEE/RSJ International Conference on Intelligent
  Robots and Systems (IROS)

\bibitem[{Wu et~al.(2020)Wu, Yan, Kurutach, Pinto, and Abbeel}]{lerrel}
Wu Y, Yan W, Kurutach T, Pinto L, Abbeel P (2020) {Learning to Manipulate
  Deformable Objects without Demonstrations}. In: Robotics: Science and Systems
  (RSS)

\bibitem[{Xie et~al.(2018)Xie, Singh, Levine, and Finn}]{fewshot_xie_2018}
Xie A, Singh A, Levine S, Finn C (2018) {Few-Shot Goal Inference for Visuomotor
  Learning and Planning}. In: Conference on Robot Learning (CoRL)

\bibitem[{Yan et~al.(2020)Yan, Vangipuram, Abbeel, and Pinto}]{yan2020learning}
Yan W, Vangipuram A, Abbeel P, Pinto L (2020) {Learning Predictive
  Representations for Deformable Objects Using Contrastive Estimation}. In:
  Conference on Robot Learning (CoRL)

\bibitem[{Yang et~al.(2017)Yang, Sasaki, Suzuki, Kase, Sugano, and
  Ogata}]{folding_2017}
Yang PC, Sasaki K, Suzuki K, Kase K, Sugano S, Ogata T (2017) {Repeatable
  Folding Task by Humanoid Robot Worker Using Deep Learning}. In: IEEE Robotics
  and Automation Letters (RA-L)

\end{thebibliography}
